\pdfoutput=1
\documentclass[11pt]{article}

\usepackage{acl}

\usepackage{times}
\usepackage{latexsym}

\usepackage[T1]{fontenc}
\usepackage[utf8]{inputenc}
\usepackage{microtype}
\usepackage{inconsolata}
\usepackage{amssymb}
\usepackage{amsthm}
\usepackage{amsmath}
\usepackage{multirow}
\usepackage{graphicx}
\usepackage{enumitem}
\usepackage{bm}
\usepackage{url}
\usepackage{hyperref}
\usepackage{todonotes}
\usepackage{subfiles}
\newtheorem{definition}{Definition}

\usepackage{caption}
\usepackage{subcaption}

\usepackage[ruled,lined,linesnumbered]{algorithm2e}
\def\BibTeX{{\rm B\kern-.05em{\sc i\kern-.025em b}\kern-.08em
    T\kern-.1667em\lower.7ex\hbox{E}\kern-.125emX}}

\newcommand{\squishlist}{ 
   \begin{list}{$\bullet$}
    { \setlength{\itemsep}{0pt}      \setlength{\parsep}{3pt} 
      \setlength{\topsep}{3pt}       \setlength{\partopsep}{0pt}
      \setlength{\leftmargin}{1.5em} \setlength{\labelwidth}{1em}
      \setlength{\labelsep}{0.5em} } }
\newcommand{\squishend}{
    \end{list}  } 

\title{Correction of Errors in Preference Ratings from Automated Metrics for Text Generation}
\author{Jan Deriu\thanks{\quad These authors contributed equally.} 
\and Pius von D\"{a}niken\footnotemark[1] 
\and Don Tuggener 
\and Mark Cieliebak \\
Centre for Artificial Intelligence\\
ZHAW School of Engineering \\
\texttt{\{deri,vode,tuge,ciel\}@zhaw.ch} \\
}

\begin{document}
\maketitle
\begin{abstract}
%A major challenge in the field of Text Generation is evaluation: Human evaluations are cost-intensive, and automated metrics often display considerable disagreements with human judgments. In this paper, we propose to apply automated metrics for Text Generation in a preference-based evaluation protocol. The protocol features a statistical model that incorporates various levels of uncertainty to account for the error-proneness of the metrics. We show that existing metrics are generally over-confident in assigning significant differences between systems. As a remedy, the model allows to combine human ratings with automated ratings. This can reduce the required amounts of human ratings, which are required to arrive at robust and statistically significant results, by more than 50\%, while yielding the same evaluation outcome as the pure human evaluation in 95\% of cases. We showcase the benefits of the evaluation protocol for three text generation tasks: dialogue systems, machine translation, and text summarization.

A major challenge in the field of Text Generation is evaluation: Human evaluations are cost-intensive, and automated metrics often display considerable disagreement with human judgments. In this paper, we propose a statistical model of Text Generation evaluation that accounts for the error-proneness of automated metrics when used to generate preference rankings between system outputs. We show that existing automated metrics are generally over-confident in assigning significant differences between systems in this setting. However, our model enables an efficient combination of human and automated ratings to remedy the error-proneness of the automated metrics. We show that using this combination, we only require about 50\% of the human annotations typically used in evaluations to arrive at robust and statistically significant results while yielding the same evaluation outcome as the pure human evaluation in 95\% of cases. We showcase the benefits of approach for three text generation tasks: dialogue systems, machine translation, and text summarization.

\end{abstract}

\section{Introduction}\label{sec:intro}
The field of Text Generation (TG) has witnessed substantial improvements over the past years. The gain in performance is mainly due to the application of large-scale pre-trained language models~\citep{devlin2019bert,reffel2020t5} based on the Transformer architecture~\cite{vaswani2017attention}, which allows fast processing of large amounts of data. This has spawned myriads of new systems for TG. The most prominent example is GPT-3~\cite{brown2020gpt3}, which showcases impressive performance on a variety of tasks in a zero-shot learning regime. %Machine translation has also improved drastically in performance, especially regarding models that handle multiple languages at once~\cite{tran2021facebook}. Similarly, the performance of conversational dialogue systems has increased substantially~\cite{adiwardana2020meena,shuster2022blenderbot}. 

One major hurdle for further progress is the evaluation of TG systems. Currently, the most reliable approach to evaluating TG systems is a human-based evaluation~\cite{celikyilmaz2020evaluation}, which is  time-consuming and cost-intensive. Furthermore, human evaluation suffers from a set of problems such as low annotator agreements~\cite{amidei2018rethinking}, and they need to be designed with care to be reproducible~\cite{belz2021reprogen}. These problems motivated the development of automated evaluation metrics, which take the input of a TG system and the generated text (and potentially one or multiple reference texts) as their input, and return a rating. 

Generally, there are two types of automated metrics: trained and untrained metrics~\cite{celikyilmaz2020evaluation}. The most prominent \emph{untrained metrics} are the BLEU score~\cite{papineni2002bleu} and the ROUGE score~\cite{lin2004rouge}, developed for the evaluation of machine translation and automated summarization systems, respectively. The more recent metrics that were proposed are \emph{trained metrics}. One of the first such approaches is the PARADISE framework by~\citet{walker1997paradise} for task oriented dialogue systems, which learns to match interaction statistics to user satisfaction scores. Current approaches are based on large pre-trained language models. For conversational dialogue systems, there are ADEM~\cite{lowe2017adem}, USR~\cite{mehri2020usr}, FED~\cite{mehri2020fed} or MAUDE~\cite{sinha2020maude}, among others (for a more complete overview, we refer the reader to~\citet{yeh2021comprehensive,deriu2021survey}). For machine translation, the most prominent trained metrics are COMET~\cite{rei2020comet} and BleuRT~\cite{sellam2020bleurt}. For a more in-depth treatment of different automated metrics for TG systems, we refer the reader to~\citet{celikyilmaz2020evaluation}.

Some metrics already achieve correlations with human judgements of 50\% and above ~\cite{yeh2021comprehensive,fabbri2021summeval,freitag2021wmt21metrics}. For this reason, it is a tempting to use automated metrics to rate and rank TG systems. A typical approach to compare two systems is to use \emph{preference ratings}, where the generated output of two systems for the same input is given, and the metric is used to decide which output is preferred or if they are of similar quality \cite{mathur-etal-2020-tangled,kocmi-etal-2021-ship}. Such preferences are then aggregated for several sample inputs to decide which system is "better". One important open question, which we will tackle in this paper, is how erroneous ratings from an automated metric on the sample level influences the system level evaluation.

\textbf{Motivating Example.} Assume that we are given a set of TG systems, and the goal is to rank them according to some criterion (e.g. relevance of generated summaries for some text summarization systems). Assume that we are given an automatic preference metric. The naive application of this metric to determine which of two TG systems is better is to apply the metric to the outputs of the two systems for a test set of a fixed size. Then one would apply a statistical significance test~\citep{sign_tests} to determine if one system is preferred significantly more often than the other by the metric. This process is repeated for each pair of systems, and then a partial ordering can be derived from the pairwise decisions. To compare the outcome of the automated evaluation, the same procedure is repeated with a human evaluation. A good metric is one that recreates the same system level preference ranking or the same pairwise results as a human evaluation would generate. 

In this setting, there are four types of outcomes with respect to a human evaluation at the system level:
\begin{itemize}[noitemsep,topsep=0pt]\label{list:error_types}
    \item \textbf{No Error}. There are two sub-cases of no errors. 1) If the human evaluation states that two systems are significantly different, and the automated evaluation states the same (Green). 2) If the human evaluation states that two systems are not significantly different, and the automated evaluation states the same (Olive).
    \item \textbf{Inversion Error}. If the human evaluation significantly prefers system A over system B, but the automated evaluation results in the opposite preference (Red).
    \item \textbf{Omission Error}. If the human evaluation states that two systems are significantly different, but the automated evaluation states that they are of the same quality (Blue).
    \item \textbf{Insertion Error}. If the human evaluation states that two systems are not significantly different, but the automated evaluation states that they are (Yellow).
\end{itemize}

We have evaluated the performance of several automated metrics in comparison to human preferences. More precisely, we examined four TG tasks  - chatbots, summarization (coherence), summarization (consistency), and machine translation - and analyzed the performance of a popular automated metric for each of these tasks when used to derive system rankings based on pairwise comparison of the metrics' sample level scores. %Figure~\ref{fig:naive_trinomial} shows a visual representation of the errors of four typical metrics for three different domains (chatbots, summarization, and machine translation). 

\begin{figure*}[!t]
\small
\centering
 \begin{subfigure}[c]{0.25\textwidth}
 \centering
     \includegraphics[width=0.5\textwidth]{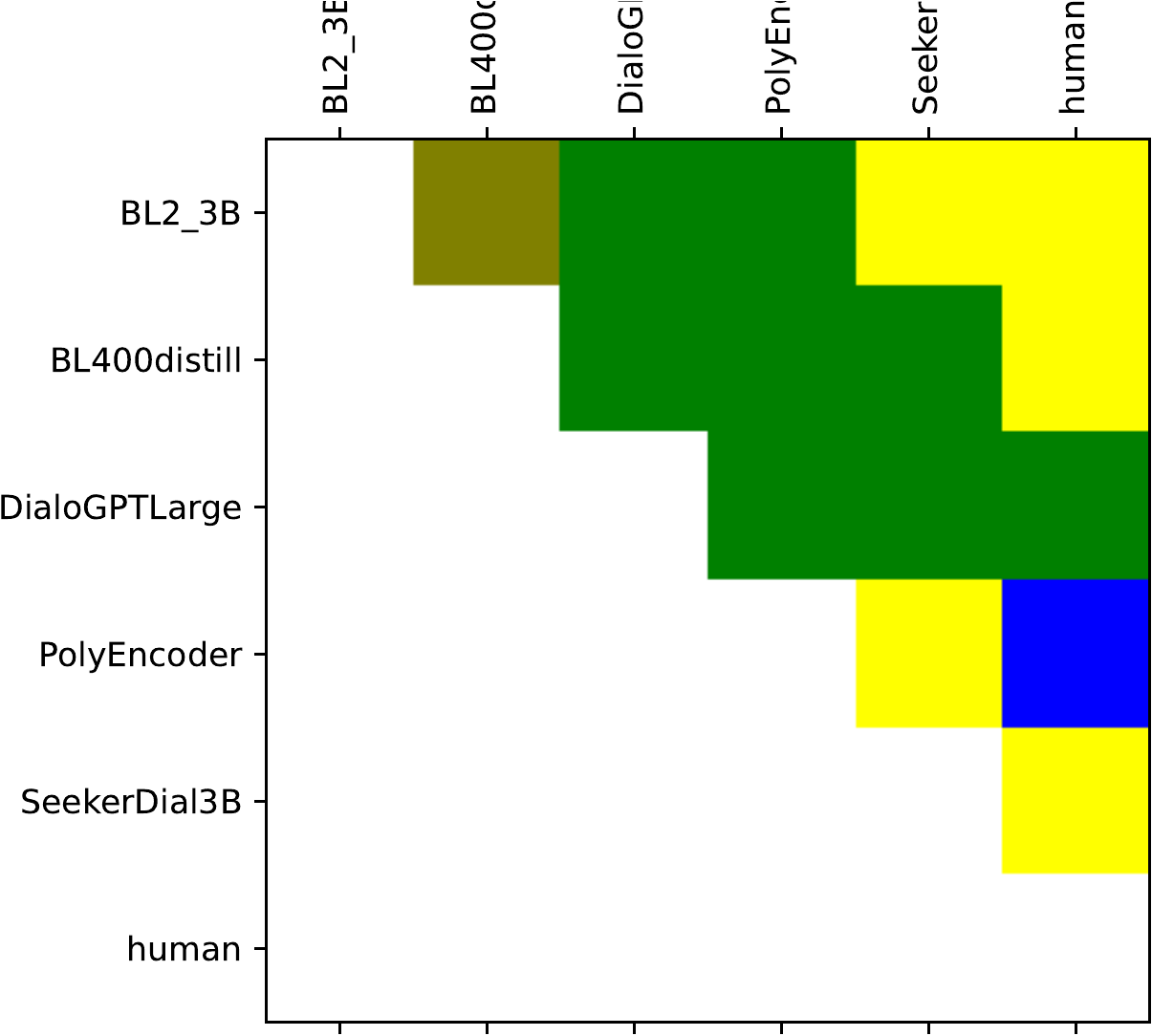}
     \caption{Chatbot-DEB}
     \label{fig:cb_deb_naive}
 \end{subfigure}
\begin{subfigure}[c]{0.25\textwidth}
 \centering
     \includegraphics[width=0.5\textwidth]{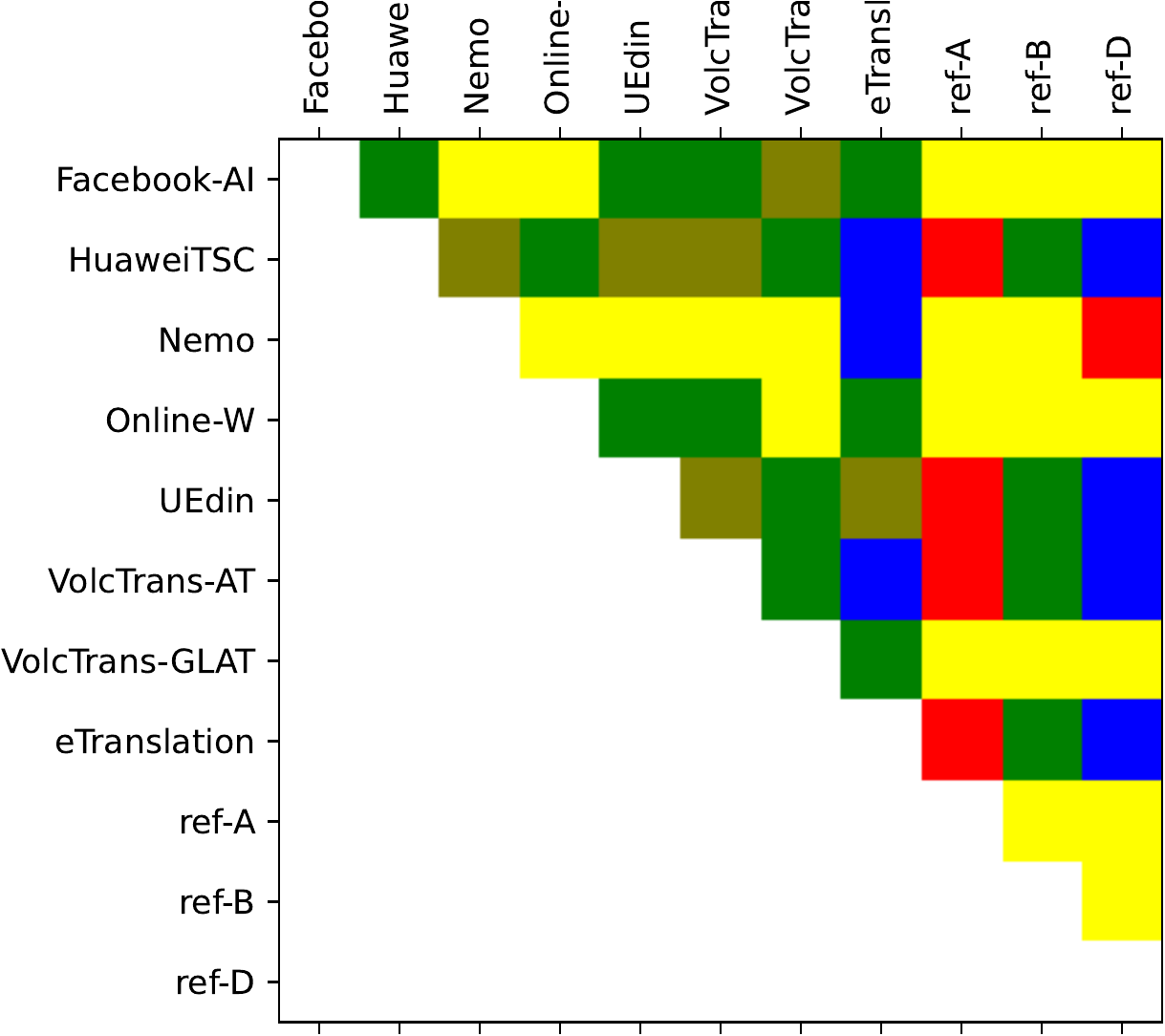}
     \caption{WMT21-COMET}
     \label{fig:wmt_comet_naive}
 \end{subfigure}
 \begin{subfigure}[c]{0.2\textwidth}
  \centering
     \includegraphics[width=0.5\textwidth]{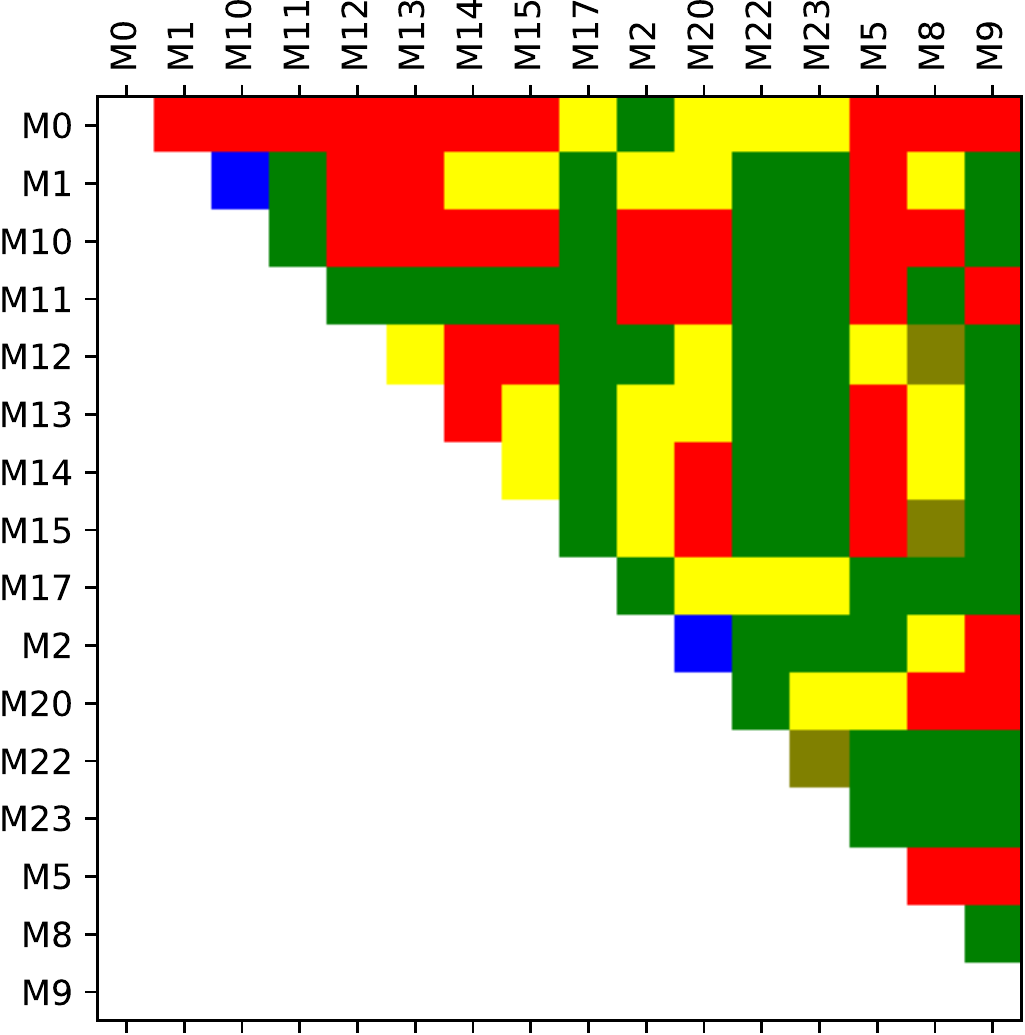}
     \caption{SummEval Coh. Rouge}
     \label{fig:summeval_coh_rouge_naive}
 \end{subfigure}
 \begin{subfigure}[c]{0.2\textwidth}
  \centering
     \includegraphics[width=0.5\textwidth]{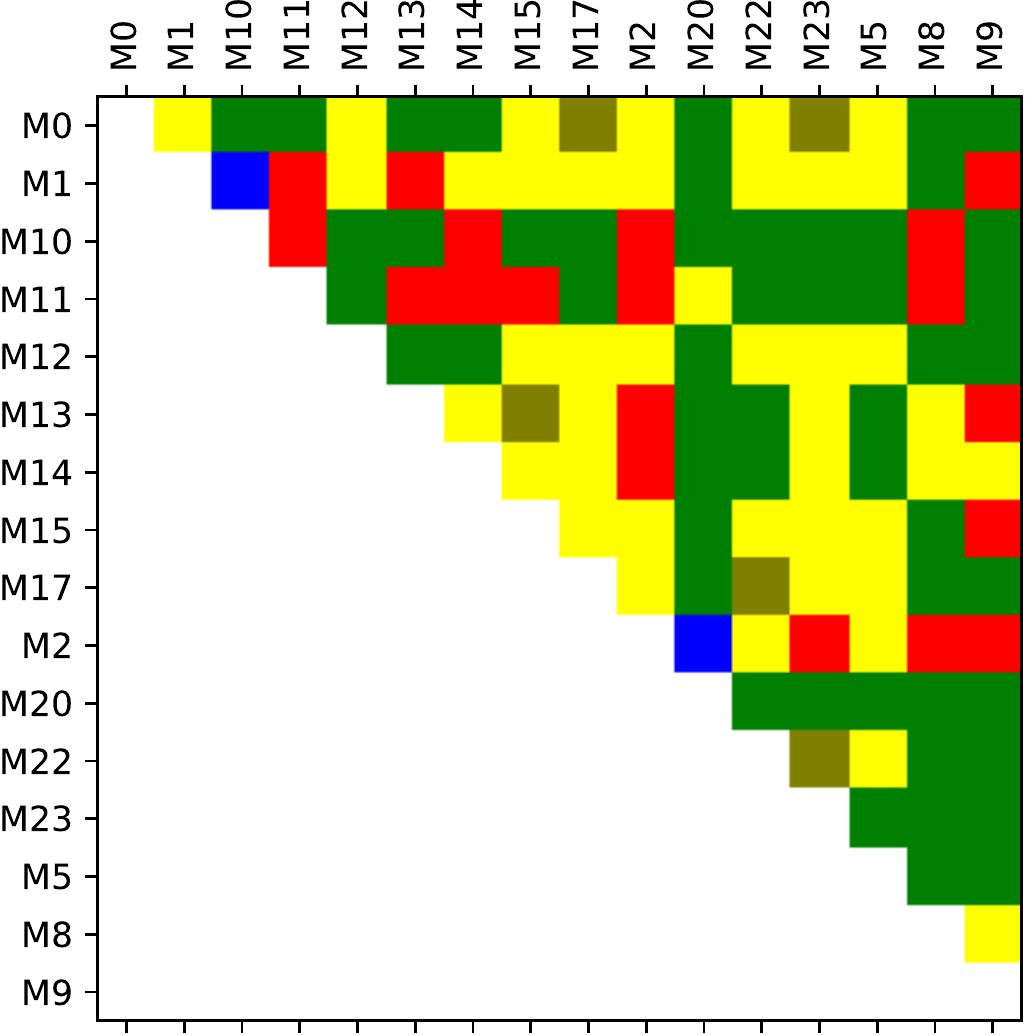}
     \caption{SummEval Cons. BertScore}
     \label{fig:summeval_cons_bert_naive}
 \end{subfigure}
    \caption{Comparison between the naive application of automated metrics and human evaluation. For each metric, the difference to the human evaluation is shown on the system-pair level. Green and Olive show agreement, Blue is an Omission Error, Yellow an Insertion Error, and Red denotes an Inversion Error.}
    \label{fig:naive_trinomial}
\end{figure*}

Figure~\ref{fig:naive_trinomial} highlights the error-proneness of the analysed automated metrics. The main findings are: 
\begin{itemize}[noitemsep,topsep=0pt]
    \item Only around 50\% of the pairwise system comparisons agree with the human evaluation.
    %\item  For all metrics, the most prominent type of error is the Insertion Error (yellow), which appears for up to 30\% of the comparisons on average.
    \item The Insertion Error is the most prominent with an average of 30\%.
    \item Inversion errors appear in around 10\% to 20\% of the comparisons on average.
    %\item We observe only a small amount of Omission errors.
    \item The are almost no Omission errors.
\end{itemize}
We hypothesize that the large discrepancy between the outcome of the automated and the human evaluation stems from three different sources of uncertainty that are not accounted for when applying the automated metric: 1) the sample size used to run the evaluation\footnote{This is sometimes addressed using standard significance testing.}, 2) the errors of the metric, and 3) the sample size used to estimate the extent of the metric errors. Thus, naively applying automated metrics leads to overconfident predictions, which yield wrong outcomes of the evaluation.

%These errors at the level of system comparison have two main sources: first, the errors that the metrics make at the sample level are not taken into account; second, the various levels of uncertainty are not accounted for, leading to overconfident predictions. These results highlight that naively applying automated metrics to derive preference ratings leads to errors in the evaluation of TG systems. In turn, this yields wrong statements about the quality and ranking of the TG systems under investigation. 

%\paragraph{Contributions.} The main contributions of this paper are twofold.  First, we propose a statistical framework that combines the human and automated evaluation and explicitly takes into account various levels of uncertainty, such as the error-proneness of the metrics and the number of samples used in the estimation process. This leads to a more robust evaluation and mends the issues highlighted in the example above. This statistical framework can be used to decide for each pair of systems if they are significantly different or if they are of the same quality. The second contribution is applying this statistical framework to devise an evaluation protocol that reduces the amount of human ratings needed to produce statistically significant results. We investigate the performance of the evaluation protocol in a case-study in three different TG tasks: chatbots, text summarization, and machine translation.  Our case studies shows that using our contributions, we can almost completely correct for the errors made by the naive application of the metrics and that the amount of human ratings needed to produce robust evaluation outcomes is reduced by more than 50\%.
\textbf{Contributions.} This paper has two main contributions: First, we propose \textit{a novel Bayesian statistical model of TG evaluation} which integrates the various sources of uncertainty mentioned above. The model yields a more robust evaluation and has the flexibility to combine human and automated evaluations. The model can be used to determine whether two systems are significantly different or if they are of equal quality. The second contribution is an \textit{evaluation protocol} that leverages the statistical model and reduces the amount of human ratings required. 

We investigate the performance of the evaluation protocol in a case-study in three different TG tasks: chatbots, text summarization, and machine translation. Our case-study shows that using our contributions, we can almost completely correct for the errors emerging in the naive application of the metrics and that the amount of human ratings needed to produce robust evaluation outcomes is reduced by more than 50\%.

\section{Definition of Preference Metrics}
In this section, we formally define preference metrics, their errors, and how to mitigate them. We then use this formalism to derive an effective evaluation protocol that can handle error-prone metrics. For the remainder of the paper, we define $\mathcal{I}$ as the set of all possible inputs (e.g., for machine translation all sentences in the source language), and $\mathcal{O}$ as the set of all possible outputs (e.g., all sentences in the target language). We start by defining a TG system as a function that takes an input and generates an output $\pi: \mathcal{I} \rightarrow \mathcal{O}$. %For instance, the input is a dialogue history and the output is the next utterance by the system. 

On an abstract level, we can define a preference metric as a function that takes as an input a triple consisting of: the input to a TG system (e.g., a sentence in the source language to be translated), the output of a system A, and the output of a system B (e.g., the translated sentences in the target language), and returns the preference rating. This is formalized as follows:

\begin{definition}[Preference Metrics]
We call functions of the form $\mathcal{M}: \mathcal{I} \times \mathcal{O} \times\mathcal{O} \rightarrow \{>, =, <\}$ \textbf{preference metrics}. We call an outcome of $">"$ a win, $"="$ a draw, and $"<"$ a loss. 
\end{definition}

Note that the semantics of $\mathcal{M}(i,o_1,o_2)$ is to find out whether output $o_1$ is preferred over $o_2$. At this point, the notion of an output being preferred is to be taken abstractly, in a real-world application this would be realized by a concrete feature (better fluency, higher relevance, etc.). 

Next, we introduce an oracle, which constitutes the ground-truth. When constructing an oracle in a real-world application, we would usually resort to human annotations. 

\begin{definition}[Preference Oracle]
    The \textbf{preference oracle} is a function 
    $\Omega : \mathcal{I} \times \mathcal{O} \times \mathcal{O} \rightarrow \{>, =, < \}$ such that
    \[    \small
        \Omega(i, o_1, o_2) =
        \begin{cases}
            > & \text{if $o_1$ is preferred to $o_2$} \\
            = & \text{if there is no preference} \\ % between $o_1$ and $o_2$} \\
            < & \text{if $o_2$ is preferred to $o_1$} \\
        \end{cases}
    \]
\end{definition}

%Now we can define the notion of an error-free preference metric, which is the metric that perfectly emulates the oracle.

%\begin{definition}[Error-Free Metric]
%    We call a comparison metric $M^{*}_{c}$ \textbf{error-free} iff it is equal to
%    the comparison oracle for all inputs and pairs of outputs:
%    \[
%        \forall i \in \mathcal{I}, o_1, o_2 \in \mathcal{O} \quad M^{*}_{c}(i, o_1, o_2) = \mathcal{O}_c(i, o_1, o_2)
%    \]
%\end{definition}

Now we define the notion of an error-prone metric. 
\begin{definition}[Error-prone Metric]\label{def:error_prone_metric}
A \textbf{preference metric with independent confusion errors} $\mathcal{M}$ is an error-prone metric where the probability of a given outcome is only dependent on the comparison oracle rating~\footnote{This means in particular that there are no dependencies on the "difficulty" of the input.}.
Its confusion probabilities are defined as:
    \begin{equation}
    {\small
        \begin{aligned}
        &\mu_{c, c'} = Pr(\mathcal{M}(i, o_1, o_2) = c | \Omega(i, o_1, o_2) = c') \\
        &\forall c, c' \in \{>, =, <\}, \forall i \in \mathcal{I},
        \forall o_1, o_2 \in \mathcal{O}
        \end{aligned}
        }
    \end{equation}
    The errors made by an error-prone preference metric can be represented by a confusion matrix with normalized columns, such that each entry in the matrix is a probability. The matrix $\bm{\mu}$ is called the \textbf{mixture matrix} of $\mathcal{M}$:

    \begin{equation}
        {\small
    \begin{aligned}
        \bm{\mu} = 
        \begin{pmatrix}
            \mu_{>>} & \mu_{>=} & \mu_{><} \\
            \mu_{=>} & \mu_{==} & \mu_{=<} \\
            \mu_{<>} & \mu_{<=} & \mu_{<<} \\
        \end{pmatrix}
        \end{aligned}}
    \end{equation}
    
\end{definition}

Note that the mixture matrix of an error-free metric is the identity matrix.

\section{Statistical Model for Preference Metrics}
\label{sec:framework}

\begin{algorithm}[!t]
\small
\caption{Pairwise Decision Function.}
\label{algo:decision}

\SetKwInOut{Input}{Input}
\SetKwInOut{Output}{Output}

\SetKwFunction{CollectAnnotations}{CollAnnotations}
\SetKwFunction{CompSignificane}{CompSignificane}
\SetKwFunction{EstimateNumAnn}{EstimateNumAnn}
\SetKwData{DecidedPairs}{\textit{DecidedPairs}}
\SetKwData{UndecidedPairs}{\textit{UndecidedPairs}}
\SetKwData{CountConfusions}{\textit{CountConfusions}}
\SetKwData{CountsFromAnn}{\textit{CountsFromAnn}}
\SetKwData{SamplePosterior}{\textit{MCMCSamplePosterior}}
\SetKwData{ComputePartialOrder}{\textit{ComputePartialOrder}}
\SetKwData{ComputeEpsBound}{\textit{ComputeEpsBound}}
\SetKwData{emptyset}{EmptySet}

\Input{$\pi_a,\pi_b$}
\Input{Set of metric judgments $M$}
\Input{Set of human judgments $A$}
\Output{Return preference rating}

\Begin{

$n_{cc'}$ $\leftarrow$ \CountConfusions($A$, $M$, c, c')

$n_>, n_=, n_< \leftarrow$ \CountsFromAnn($A$)

$m_>, m_=, m_< \leftarrow$ \CountsFromAnn($M$)

$\bm{\cdot\mu}_> \sim Dir(n_{>>} +1 , n_{=>} +1,n_{<>} +1)$

$\bm{\cdot\mu}_= \sim Dir(n_{>=} +1 , n_{==} +1,n_{<=} +1)$

$\bm{\cdot\mu}_< \sim Dir(n_{><} +1 , n_{=<} +1,n_{<<} +1)$

$\bm{\mu} = (\bm{\mu_{\cdot>}}, \bm{\mu_{\cdot=}}, \bm{\mu_{\cdot<}})$

$\bm{p} \sim Dir(n_> + 1, n_= + 1, n_< + 1 )$

$m_>, m_=, m_< | \bm{p}, \bm{\mu}  \sim Tri(|M_{ij}|, \bm{\mu}\bm{p})$

$\left\{ \tilde{\bm{p}} \right\}_{i = 1}^{N} \leftarrow \SamplePosterior(\bm{p}, N)$

$\theta = \frac{1}{N}\sum_{i = 1}^{N} \mathbb{I}[\tilde{p_i}_> > \tilde{p_i}_<]$

\uIf{$\theta > 1 - \frac{\gamma}{2}$}{
    \Return >
}
\uElseIf{$\theta < \frac{\gamma}{2}$}{
    \Return <
}
\Else{
    \Return =
}

}
\end{algorithm}

In this section, we introduce the statistical model that is used to compare two TG systems $\pi_a$ and $\pi_b$. The model encompasses three main sources of uncertainty. 
\begin{enumerate}[noitemsep,topsep=0pt]
%\item Sample size for the system comparison: Are there enough samples to assess a significant difference between the systems?
\item Uncertainty due to sample size of both error-free and error-prone ratings
%\item Metric errors: What is the quality of the metrics and how do the metric's errors impact the comparison between $\pi_a$ and $\pi_b$?
\item Uncertainty introduced by errors from the error-prone metric
%\item Sample size for assessing the metrics errors: Are there enough oracle samples to reliably estimate the metric's error rates?
\item Uncertainty over the true error-rates of the error-prone metric
\end{enumerate}
We build up the statistical model step-by-step by discussing each source of uncertainty. We apply the Bayesian approach which allows us to describe the process in terms of probability distributions that can be sampled by using Markov Chain Monte Carlo (MCMC) sampling (refer to Appendix~\ref{sec:derivations} and~\ref{sec:mcmc_samp} for additional details). 

For the rest of the chapter, assume that we have access to a set of inputs $\{i_1, \ldots, i_n\} \subseteq \mathcal{I}$, and the corresponding system outputs of $\pi_a$ and $\pi_b$, i.e., $o_j^a = \pi_a(i_j)$, and $o_j^b = \pi_b(i_j)$.

\subsection{Step One: Direct Estimation of the Win-Rate Significance}
For this, assume that we have access to the preference oracle $\Omega$ itself. Let $r_j^{\Omega} = \Omega(i_j, o_j^a, o_j^b)$ be the output of the preference oracle. Let $\mathbb{I}^{x}[y]$ denote an index function that is equal to $\mathbb{I}^{x}[y] = 1 \iff x=y$, and $0$ otherwise. Then let $n_{>} = \sum_{j=1}^{n} \mathbb{I}^{>}[r_j^{\Omega}]$ denote the number of times $o_j^a$ was rated as being better than $o_j^b$. We analogously define  $n_{<} = \sum_{j=1}^{n} \mathbb{I}^{<}[r_j^{\Omega}]$ the number of times $o_j^a$ was rated as being worse than $o_j^b$, and $n_{=} = \sum_{j=1}^{n} \mathbb{I}^{=}[r_j^{\Omega}]$ the number of draws. 
We use a Dirichlet distribution to model the posterior distribution for the given observations: 
\begin{equation}
\small
\begin{aligned}
&Pr(\bm{p} | N_> = n_>, N_= = n_=, N_< = n_<) \\ 
& \sim Dir(n_> + 1, n_= + 1, n_< + 1)
\end{aligned}\label{eq:oracle_posterior}
\end{equation}
where $\bm{p} = (p_>, p_=, p_<)$ denotes the probability vector for the win-rate $p_>$, the draw-rate $p_=$, and the loss-rate $p_<$.

\subsection{Step Two: Integrate the Metric-Errors}\label{sec:step_two}
In this section, we assume a metric that makes mistakes, i.e., an error-prone metric $\mathcal{M}$. Let $r_j^m = \mathcal{M}(i_j, o_j^a, o_j^b)$ denote the rating given by an error-prone metric for sample j. Analogous to before, we define $m_{>} = \sum_{j=1}^{n} \mathbb{I}^{>}[r_j^m]$ the number of times the error-prone metric prefers the output of system $\pi_a$ to that of $\pi_b$. The counts for equality and being worse are denoted by $m_=$ and $m_<$, respectively. Since $\mathcal{M}$ is an error-prone metric, the counts $m_{>,=,<}$ are not equal to the true counts $n_{>,=,<}$, which are yielded by an oracle. The errors made by the metric are characterized by its mixture matrix $\bm{\mu}$. In this section, we assume that the precise values of $\bm{\mu}$ are known. We note that the true probabilities $\bm{p} = (p_>, p_=, p_<)$ are transformed by the mixture matrix to the error-prone ones by $\hat{\bm{p}} = (\hat{p}_>, \hat{p}_=, \hat{p}_<) = \bm{\mu} \bm{p}$. We want to model the posterior distribution $p(\bm{p} | M_> = m_>, M_= = m_=, M_< = m_<)$ of the true probabilities $\bm{p}$ given the observed error-prone ratings. This is done by combining the prior belief of $\bm{p}$ with the likelihood of the observed $m_{>, =, <}$ values, which can be modeled using a Multinomial distribution. We use a Dirichlet prior for $\bm{p} \sim Dirichlet(\alpha_>, \alpha_=, \alpha_<)$. The parameters $\alpha_c$ are either chosen according to Equation \ref{eq:oracle_posterior}, if we have access to oracle ratings, or set to $1$, which corresponds to a uniform prior. 
\begin{equation*}\label{eq:fixed_mu}
{\small
    \begin{aligned}
        &\bm{p} \sim Dirichlet(\alpha_>, \alpha_=, \alpha_<) \\
        &m_>, m_=, m_< | \bm{p} \sim Mult(n, \bm{\mu}\bm{p}) \\
        &Pr(\bm{p} | m_>, m_=, m_<)\\ 
        &\propto Pr(M_> = m_>, M_= = m_=, M_< = m_< | \bm{p})Pr(\bm{p}) 
    \end{aligned}
    }
\end{equation*}
%We can apply Markov Chain Monte Carlo (MCMC) sampling to produce samples from the resulting posterior. We again compute the fraction of samples where $p_> > p_<$ to compute whether the comparison is significant at the $\gamma$ level.

\subsection{Step Three: Integrate Uncertainty over Error Measurements}\label{sec:step_three}
In a real-world scenario, the values of $\bm{\mu}$ must be estimated from data. This is achieved by comparing the error-prone metric outputs to a set of oracle outputs. For this, we use the following counts: $n_{c,c'} = \sum_{i = 1}^{n} \mathbb{I}^c[r_j^m] * \mathbb{I}^{c'}[r_j^{\Omega}] ,\forall c, c' \in \{>, =, <\}$. Thus, $n_{<,=}$ denotes the number of times the error-prone metric returns $<$ and the oracle returns $=$. In Bayesian terms, each column in the mixture matrix is modeled as a Dirichlet distribution.
\begin{equation}\label{eq:mu:diriclet}
{\small
    \begin{aligned}
        &\bm{\mu_{\cdot>}} = (\mu_{>>}, \mu_{=>}, \mu_{<>})^\top \sim Dirichlet(n_{.,>} + 1) \\
        &\bm{\mu_{\cdot=}} = (\mu_{>=}, \mu_{==}, \mu_{<=})^\top  \sim Dirichlet(n_{.,=} + 1) \\
        &\bm{\mu_{\cdot<}} = (\mu_{><}, \mu_{=<}, \mu_{<<})^\top \sim Dirichlet(n_{.,<} + 1) \\
        &\bm{\mu} = (\bm{\mu_{\cdot>}}, \bm{\mu_{\cdot=}}, \bm{\mu_{\cdot<}}) \\
    \end{aligned}
    }
\end{equation}
Thus, the mixture matrix is treated as a random variable. Putting all together, we define a joint posterior for $\bm{p}$ and $\bm{\mu}$ given the error-prone metric observation and the prior for $\bm{p}$ and $\bm{\mu}$.
\begin{equation*}\label{eq:comp_full_bayes}
\small
    \begin{aligned}
        & \bm{p} \sim Dirichlet(\alpha_>, \alpha_=, \alpha_<) \\
        & \bm{\mu} \sim \text{see Equation~\ref{eq:mu:diriclet}} \\
        & m_>, m_=, m_< | \bm{p}, \bm{\mu} \sim Mult(n, \bm{\mu}\bm{p}) \\
        & Pr(\bm{p}, \bm{\mu} | m_>, m_=, m_<)  \\
        & \propto Pr(m_>, m_=, m_< | \bm{p}, \bm{\mu})Pr(\bm{p})Pr(\bm{\mu}) \\
    \end{aligned}
\end{equation*}
%To determine whether $p_g > p_l$ is significant, we apply MCMC sampling in the same way as above.

\subsection{Decision Function}
Algorithm \ref{algo:decision} shows how to apply the framework for one pair of systems $\pi_a, \pi_b$ for a set of inputs $I$ and a set of human annotations $A$. First the metric $\mathcal{M}$ is used to generate the set of automated ratings $M$. Then the confusion counts $n_{c,c'}$ are computed based on the human annotations and the metric ratings, which are used to create the distributions of the mixture matrix $\bm{\mu}$. Then the human annotations are used to estimate the prior distribution $Pr(\bm{p})$ of the comparison results. The metric samples are then used to estimate the posterior distribution $Pr(\bm{p}, \bm{\mu} | m_>, m_=, m_<)$. Each of the three steps presented above yields a posterior distribution for $\bm{p} = (p_>, p_=, p_<)$. In order to decide whether system $\pi_a$ is better than system $\pi_b$, we need to check whether $p_>$ and $p_<$ are significantly different. For this, we draw a number of samples $\tilde{\bm{p}_i}$ from the posterior. This can in general be done using Markov Chain Monte Carlo sampling~\footnote{The posterior in Equation~\ref{eq:oracle_posterior} can be sampled directly.}. We define a significance level $\gamma$ (e.g. $\gamma = 0.05$) and consider the fraction of samples where $\tilde{p_i}_> > \tilde{p_i}_<$. If this fraction is greater than $1 - \frac{\gamma}{2}$, then we regard the difference as being significant. Conversely, if the fraction is smaller than $\frac{\gamma}{2}$, then $\pi_a$ is significantly worse than $\pi_b$.

%The decision function is based on whether the probability that system $\pi_a$ is better than system $\pi_b$ is higher than the probability of the other way around, that is whether $p_> > p_<$. In order to determine whether $p_> > p_<$ is significant, we can directly sample from the posterior in Equation \ref{eq:oracle_posterior}. We define a significance level $\gamma$ (e.g. $\gamma = 0.05$) and consider the fraction of samples where $p_> > p_<$. If this fraction is greater than $1 - \frac{\gamma}{2}$, then we regard the difference as being significant. Conversely, if the fraction is smaller than $\frac{\gamma}{2}$, then $\pi_a$ is significantly worse than $\pi_b$. 

%In order to determine whether $p_> > p_<$ is significant, we apply MCMC sampling to the posterior distribution. If $p_> > p_<$ occurs in more than $1 - \frac{\gamma}{2}$ percent of cases, then we regard the difference as being significant. Conversely, if $p_> > p_<$ occurs in only less than $\frac{\gamma}{2}$ percent of cases, then $\pi_a$ is significantly worse than $\pi_b$. 

%Using Markov Chain Monte Carlo to sample from the posterior distribution, we sample compute the probability $p_g$, and check the percentage of cases where it is larger than $p_l$, if this occurs more then $1 - \gamma$ percent of times, we state that $\pi_a$ is significantly better than $\pi_b$.  

\section{Evaluation Protocol}
\begin{algorithm}[!t]
\small
\caption{Evaluation Protocol} %using Error-prone Metrics}
\label{algo:eval_prot}

\SetKwInOut{Input}{Input}
\SetKwInOut{Output}{Output}

\SetKwFunction{CollectAnnotations}{CollAnn}
\SetKwFunction{CompSignificane}{CompSignificane}
\SetKwFunction{EstimateNumAnn}{EstimateNumAnn}
\SetKwData{DecidedPairs}{\textit{DecidedPairs}}
\SetKwData{UndecidedPairs}{\textit{UndecidedPairs}}
\SetKwData{ComputeMixture}{\textit{ComputeMixture}}
\SetKwData{CountsFromAnn}{\textit{CountsFromAnn}}
\SetKwData{SamplePosterior}{\textit{MCMCSamplePosterior}}
\SetKwData{ComputePartialOrder}{\textit{ComputePartialOrder}}
\SetKwData{ComputeEpsBound}{\textit{ComputeEpsBound}}
\SetKwData{DecisionFunction}{\textit{DecisionFunction}}
\SetKwData{emptyset}{EmptySet}

\Input{$\Pi = \pi_1,...\pi_S$}
\Input{An error-prone automated metric $\mathcal{M}$}
\Input{Budget for human annotations $B$}
\Input{Annotation batch size $N$}
\Input{Set of inputs $I$}
\Output{Partial Order of $\pi_1,...\pi_S$}

\Begin{

\UndecidedPairs $\leftarrow$ $\forall i,j: (\pi_i, \pi_j) \in \Pi \times \Pi$;

$A_{ij}$ $\leftarrow$ \emptyset\;
$M_{ij} = \{\mathcal{M}(i_k, \pi_i(i_k), \pi_j(i_k)) | \forall k \in |I|\}$

\While{$B > 0$ and $|\UndecidedPairs| > 0$}{
    \ForEach{$\pi_i, \pi_j \in$ \UndecidedPairs}{
        $A_{ij} \leftarrow A_{ij} \cup$ \CollectAnnotations($N$, $\pi_i$, $\pi_j$)
        $B \leftarrow B - N$
    }
    \ForEach{$\pi_i, \pi_j \in$ \UndecidedPairs}{
        op = \DecisionFunction($A_{ij}$, $M_{ij}$, $\pi_i$, $\pi_j$)
        
        \If{op $\in \{<, >\}$}{
            \UndecidedPairs $\leftarrow$ \UndecidedPairs $\setminus (\pi_i, \pi_j)$
        }
    }
}
\Return \ComputePartialOrder($\Pi \times \Pi \setminus$ \UndecidedPairs);
}
\end{algorithm}
%In this section, we present an evaluation protocol that reduces the number of human annotations that are required to determine which TG systems are better than others.
In this section, we present an evaluation protocol combining human and automated metric ratings assuming a limited budget for human annotations.
More formally, given a set of TG systems $\pi_1,...,\pi_S$, we want to create a partial order, where $\pi_i > \pi_j$ if the win-rate of $\pi_i$ is significantly greater than the one from $\pi_j$. The evaluation protocol is depicted in Algorithm \ref{algo:eval_prot}. The protocol leverages the statistical framework to reduce the amount of human annotations needed by leveraging the metric judgments. This works as follows: We are given a set of inputs $I$ (which corresponds to a test set), a set of TG systems $\{\pi_1,...,\pi_S\}$ to be ranked, an automated metric $\mathcal{M}$, and an annotation budget $B$, which is the maximum allowed amount of annotations. The result of the protocol is a (potentially) partial order of the TG systems.

The protocol starts with a set of undecided system pairs, which initially consists of all pairs of systems, and an empty set of human annotations for each pair of systems $A_{ij}$. In a first step, the metric $\mathcal{M}$ computes the scores $M_{ij}$ for each pair of systems. That is, for all inputs in $I$ all TG systems generate their outputs, which are then evaluated using $\mathcal{M}$. 

Then we repeat the following process until our budget is empty. First we extend $A_{ij}$ with a batch of $N$ human annotations for each pair of undecided systems. We then iterate over the undecided system pairs and use the decision function from Algorithm \ref{algo:decision} (see Section~\ref{sec:framework}) to decide whether two given systems are significantly different given the current set of annotations and metric ratings. If so, the pair is removed from the set of undecided pairs. When the budget is empty or all system pairs are decided, a (potentially) partial order is computed. The decision function leverages human and automated ratings to state whether one system is significantly better than the other.

The advantage the protocol is two-fold. First, it exploits the fact that some system pairs are easier to distinguish than others. In cases where $|p_> - p_<|$ is large we need fewer human annotations to achieve the significance threshold. Compared to the setting where we allocate the same number of human annotations for each pair of systems, this allows us to spend more of the annotation budget on difficult system pairs. This approach can be used even in the absence of automated ratings. Second, our framework allows for a seamless combination of ratings from both humans and an automated metric.

%The advantage of the protocol is that is leverages the decision function, which uses both human and automated ratings to yield more trustworthy and robust decisions. With this, less human annotations are needed, as automated ratings are now leveraged soundly. 

\section{Case Studies - Setup}
In this section, we present three case studies where we apply the evaluation protocol outlined in Algorithm~\ref{algo:eval_prot}. As showcases, we use three domains: the WMT 21 metrics task data \cite{freitag2021wmt21metrics} for machine translation, the SummEval data \cite{fabbri2021summeval} for summarization, and data collected  for conversational dialogue systems (see Appendix \ref{sec:cs_details}). Table \ref{tbl:data} gives an overview of the setting. For each domain, we investigate a set of metrics applied to outputs of a set of TG systems. We provide the details of the TG systems and the metrics in Appendix \ref{sec:cs_details}. 

\begin{table}[t!]
    \small
    \begin{tabular}{ l | c | c | c }
&Chatbot&SummEval&WMT21 \\\hline
Data&BST&CNN/DM&News EN->DE \\
Metrics & 5 & 7 & 4 \\
TG Systems & 6 & 11 & 16 \\
Human Ratings & 50 & 100 & 500 \\
Metric Ratings & 1k & 11k & 1k
\end{tabular}
    \caption{Overview of the data used. The ratings refer to the number of ratings available for each pair of TG systems.}
    \label{tbl:data}
\end{table}

\noindent{\bf Chatbot:} For the chatbot domain, we used the ParlAI framework~\cite{miller2017parlai} to generate 1000 outputs for 5 different TG systems on the BlendedSkillTask (BST) dataset~\cite{smith-etal-2020-BST}. We then used the DialEval framework by~\newcite{yeh2021comprehensive} to run the outputs on 5 different metrics: DEB~\cite{sai2020deb}, GRADE~\cite{huang2020-grade}, HolisticEval~\cite{pang2020towards}, MAUDE~\cite{sinha2020maude}, and USL-H~\cite{phy2020deconstruct}. In addition, we  used Amazon Mechanical Turk to annotate 50 pairwise outputs.  \\
\noindent{\bf SummEval:} For the summarization domain, we used the SummEval framework~\cite{fabbri2021summeval}, which provides the outputs of 16 different summarization tools on the CNN/DailyMail corpus~\cite{nallapati2016abstractive}, as well as 100 expert annotations for each of these systems for each of the features: relevance, coherence, consistency, and fluency. We used the SummEval framework to generate 11k pairwise ratings by 7 different automated metrics: BertScore~\cite{zhang2019bertscore}, BLANC~\cite{vasilyev2020blanc}, CIDEr~\cite{Vedantam2015cider}, Rouge-L~\cite{lin2004rouge}, S3~\cite{peyrard2017s3}, SummaQA~\cite{scialom2019summaqa}, and SUPERT~\cite{gao2020supert}. \\
\noindent{\bf WMT21:} For machine translation, we used the WMT21 metrics task data~\cite{freitag2021wmt21metrics}. In this work, we only focus on the English to German language pair and the news domain, where eight machine translation systems were evaluated, plus three human references for each input which were also regarded as TG systems  (resulting in eleven TG systems). Although the WMT21 metrics task inspected 15 different automated metrics, we only focused on four of them (we selected the most prominent ones): BleuRT~\cite{sellam2020bleurt}, COMET~\cite{rei2020comet}, C-SPEC~\cite{takahashi2021cspec}, and sentence-level BLEU~\cite{papineni2002bleu}. For each TG system there are 500 expert MQM annotations, and for each metric there are 1000 metric ratings.

\section{Case Studies - Results}
In this section, we discuss the results of the case study. We use the error-measures that we presented in List \ref{list:error_types} in the introduction.
Furthermore, for each system pair we compute the Kullback-Leibler Divergence (KLD) between the mode of the posterior in Equation~\ref{eq:oracle_posterior} based on all human annotations $\bm{p}_{hum}$ and the mean estimated by running Algorithm~\ref{algo:eval_prot} $\bm{p}_{prot}$. We then report the average over all pairs of systems: $\frac{2}{S(S-1)}\sum_{j > i} KLD(\bm{p_{prot}}^{(i,j)} ||\bm{p_{hum}}^{(i,j)})$. 
Note that in Tables \ref{tbl:naive_res} and \ref{tbl:protocol_res}, we only report the Relevance part of the SummEval data due to space limitations. The results for the other features are in Appendix \ref{sec:full_res}.

\begin{table}[t!]
\centering
\small
\begin{tabular}{ l | c c c c | c}
Domain & Corr. & Inv. & Omi.. & Ins. & KLD \\\hline 
Chatbot             & 0.47 & 0.1 & 0.10 & 0.33 & 0.46 \\
Summeval Rel. & 0.55 & 0.19 & 0.03 & 0.23 & 0.28 \\
WMT21               & 0.52 & 0.07 & 0.10 & 0.31 & 0.52 \\ \hline 
\end{tabular}

\caption{Average frequency of Error Types for all metrics if metrics are applied naively. Correct (Cor.), Inverted (Inv.), Omission (Omi.), and Insertion (Ins.).}
\label{tbl:naive_res} 
\end{table}

\textbf{The naive application of metrics yields many errors.} When applying the metrics naively, i.e. by simply checking whether $m_>$ is significantly bigger than $m_<$, then this introduces many cases where systems that are not significantly different are rated as such. 
%In Figure \ref{fig:naive_trinomial} the outcomes are depicted for four exemplar cases. In all examples, there are many cases of Insertion Errors (yellow). For the SummEval and WMT21 case, there are also cases of Inversion Errors (red). These results are explained by two facts. First, since in this naive usage of metrics, the uncertainty from their errors are not included, many pairs of systems are rated as being significantly different even if they are not. Second, since the errors of metrics are not symmetric, that is, they are often biased in favor or against certain systems, the results tend to have many inversion errors. For instance, in WMT21 the COMET metric is biased against the ref-A, which leads it to state that it is worse than some machine translation system, although the full human evaluation comes to the opposite conclusion. 
In Table~\ref{tbl:naive_res} the average error rates for each error type are shown (the full result table is found in Appendix~\ref{sec:full_res}). Overall the Insertion error type dominates (averaging at 23\% to 33\%). That is, in all domains, the metrics have a strong tendency to suggest differences between systems that are not statistically significant according to the human evaluation. The rate of inversion errors depends on the domain and metrics used. For the chatbot domain, the average Inversion error rate lies at 10\%. For SummEval-Relevance the Inversion error rates lie at an average of 23\%. The average KLD scores are high, which indicates that the naive application of metrics yields distributions that are in high disagreement with the human evaluation. 
%We note that the ideal metric (Ideal-$\mathcal{M}$) only makes Insertion errors, which is due to the fact that only 50\% of draws are correctly predicted. However, due to the high accuracy in detecting wins and losses, there are no inversion or omission errors. This is also reflected with the small KLD score.

%\noindent\textbf{Correction without human ratings yields omission errors.} When applying the correction to metrics without using the human data as a prior (i.e., $\bm{p} \sim Dirichlet(1,1,1) \sim Uniform(0,1)$), then most metrics yield no significant differences between TG systems. All the errors are Omission Errors, and the correct decisions are mostly assigning no significant difference. The reason for this is two-fold: first, not using the human ratings as prior information introduces uncertainty, and second, the quality of the metrics is not yet good enough to correctly measure the difference between two TG systems. These results showcase the utility of our correction approach, since it models the various sources of uncertainty, which avoids that metrics make wrong claims (Inversion and Insertion Errors). %Furthermore, modeling the different sources of uncertainty showcases the demand for higher-quality metrics. 

\begin{table}[t!]
\centering
\small
\begin{tabular}{ l | c c c c | c | c}
Metric & Corr. & Inv. & Omi. & Ins. & KLD & Ann. \\ \hline
\multicolumn{ 7 }{c}{Chatbot Domain} \\ \hline
Human    & 0.93 & 0.00 & 0.00 & 0.07 & 0.05 & 0.59\\
DEB      & 0.93 & 0.00 & 0.00 & 0.07 & 0.06 & 0.49 \\
GRADE    & 0.87 & 0.00 & 0.00 & 0.13 & 0.05 & 0.53 \\
HOLISTIC & 0.87 & 0.00 & 0.00 & 0.13 & 0.05 & 0.52 \\
MAUDE    & 0.87 & 0.00 & 0.00 & 0.13 & 0.05 & 0.53 \\
USL-H    & 0.87 & 0.00 & 0.00 & 0.13 & 0.05 & 0.52 \\ \hline \hline
\multicolumn{ 7 }{c}{Summeval Relevance Domain} \\ \hline
Human     & 0.96 & 0.00 & 0.00 & 0.04 & 0.07 & 0.43 \\
BertScore & 0.90 & 0.00 & 0.01 & 0.09 & 0.06 & 0.40 \\
BLANC     & 0.93 & 0.00 & 0.02 & 0.06 & 0.05 & 0.41 \\
CIDEr     & 0.93 & 0.00 & 0.01 & 0.06 & 0.06 & 0.42 \\
ROUGE-L   & 0.92 & 0.00 & 0.01 & 0.08 & 0.05 & 0.41 \\
S3        & 0.93 & 0.00 & 0.01 & 0.07 & 0.06 & 0.41 \\
SummaQA   & 0.92 & 0.00 & 0.01 & 0.08 & 0.05 & 0.41 \\
SUPERT    & 0.91 & 0.00 & 0.01 & 0.08 & 0.06 & 0.39 \\ \hline \hline
%Ideal-$\mathcal{M}$ & 0.85 & 0.00 & 0.00 & 0.15 & 0.02 & 0.38 \\ \hline \hline
\multicolumn{ 7 }{c}{WMT21 Domain} \\ \hline
Human      & 0.65 & 0.00 & 0.00 & 0.35 & 0.06 & 0.34\\
BleuRT     & 0.65 & 0.00 & 0.00 & 0.35 & 0.06 & 0.34 \\
C-SPEC     & 0.65 & 0.00 & 0.00 & 0.35 & 0.06 & 0.33 \\
COMET      & 0.65 & 0.00 & 0.00 & 0.35 & 0.06 & 0.34 \\
BLEU       & 0.65 & 0.00 & 0.00 & 0.35 & 0.05 & 0.34 \\ \hline
\end{tabular}

\caption{Frequency of Error Types for all metrics if the protocol is applied. Correct (Cor.), Inverted (Inv.), Omission (Omi.), Insertion (Ins.), and the fraction of annotations needed with the protocol (Ann.).}
\label{tbl:protocol_res} 
\end{table}

\noindent\textbf{The evaluation protocol is able to recreate the original results.} Table~\ref{tbl:protocol_res} shows the results of applying the protocol described in Algorithm~\ref{algo:eval_prot}. We also report the results achieved when applying the protocol using only human ratings (i.e., leaving $M_{ij}$ empty), as well as the result of an ideal metric in the SummEval - Relevance case. 
First, we note that there are no Inversion Errors, almost no Omission Errors, and there is a high Correctness score. For the Chatbot and SummEval domain the outcomes agree in around 90\% to those of the human evaluation. For the WTM21 domain, the agreement is lower at 65\%. The most common error type is the Insertion Error. In our setting this can be explained by the fact that we are using the outcomes of significance tests to compare the human to the protocol evaluation. Thus, using corrected metric samples increases the amount of samples, which leads to pairs being rated as significantly different. Since the ratings are based on our decision function, which takes into account different sources of uncertainty, the Insertions are not necessarily wrong. In fact one reason to use automated evaluation is to find differences between system that would be too expensive to discover with human annotations. 

A different view for comparing the outcomes of the evaluations is given by the KLD score, which reports how close the distribution $\bm{p}_{prot}$ is to the original human evaluation. This view removes the significance test from the equation, and better showcases the disagreement between the protocol and the original human evaluation. In all cases the KLD scores are very low, which shows that the protocol yields results comparable to the original human evaluation.
\begin{figure}[!t]
\small
\centering
\includegraphics[width=0.452\textwidth]{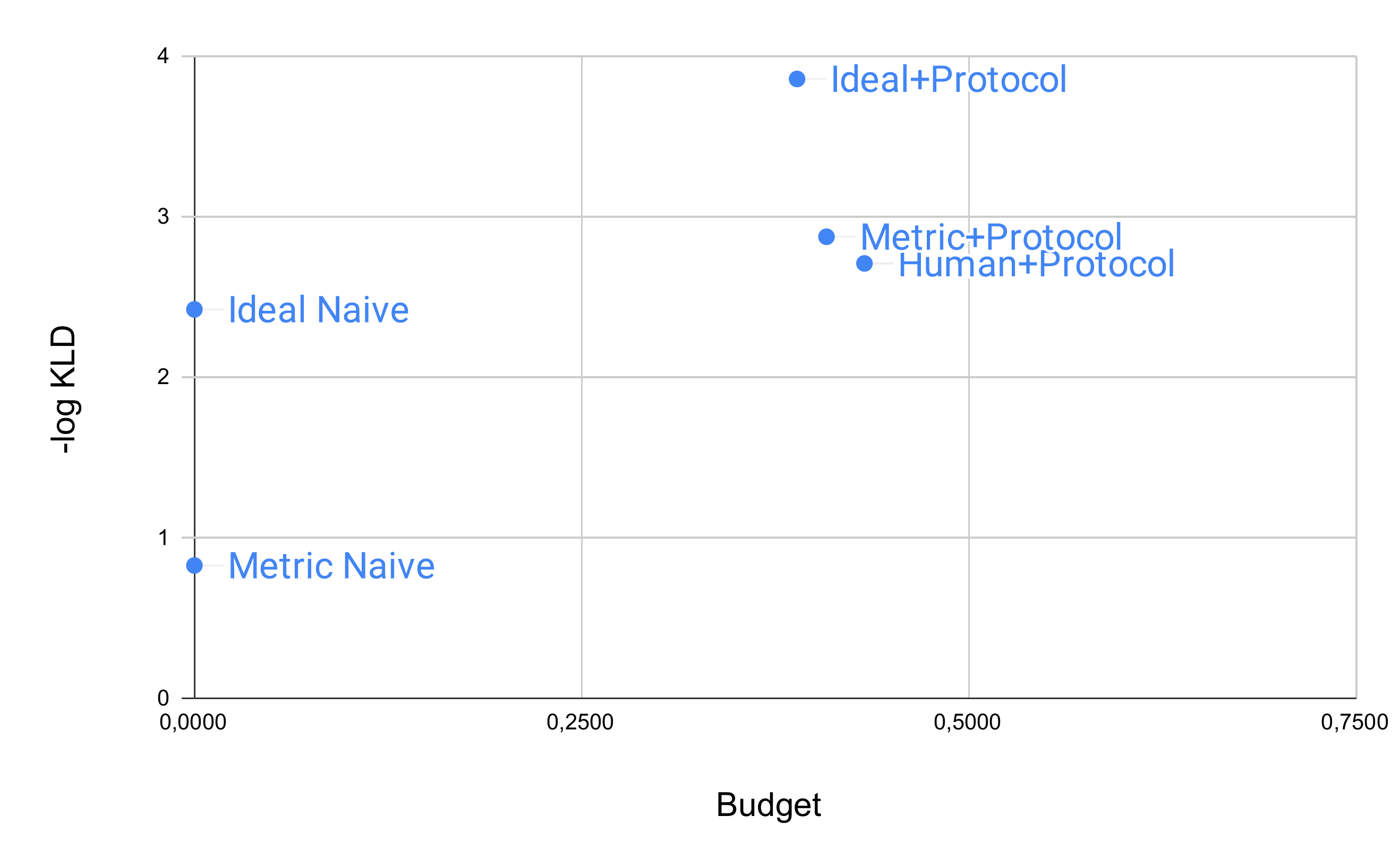}
\caption{Budget vs. Kullback-Leibler Divergence}
\label{fig:budgetkld} 
\end{figure}
In terms of the number of annotations needed, there are two measures. First, we compare the number of annotations needed by the protocol to the one needed by the full human evaluation. 
Here, the application of our protocol reduces the amount of humans annotation by more than half in most cases. For the WMT21 we can even reduce it by two thirds.
The second view is comparing the number of annotations needed by the protocol to the annotations needed when the protocol is applied to human ratings only. For the Chatbot and SummEval domain, leveraging automated metrics results in less data needed (up to 10\% for the Chatbot domain, and 5\% for the SummEval domain). For the WMT21 domain, only 1\% difference is measured. We assume that this are due to the fact that the metrics are not yet of high enough quality to yield the boost needed to have a large impact. \\
\noindent\textbf{Summary.} Figure \ref{fig:budgetkld} summarizes the main outcomes of this work. The Figure shows the number of annotations (x-axis) in relation to the negative log-KLD score achieved (y-axis). The full human evaluation is set as reference, that is, using 100\% of annotations, and a KLD score of 0 (thus, not shown). The Figure shows that on average using the full protocol on real-world metrics yields using 40\% of annotations, and achieving a KLD score of 0.08. On the other hand, not using metric ratings in the protocol needs 43\% of annotations and achieving a worse KLD score of 0.6. The naive application of the metrics does not need any annotations but yields high KLD scores (1.6 on average).  To showcase an upper limit, we also added the KLD divergence for an ideal metric, which we simulated using the Bayesian model, where we use a fixed $\bm{\mu}$ (see Appendix~\ref{sec:ideal_met} for details). The ideal metric only needs 38\% of annotation, and achieves a KLD score of 0.02. An ideal metric would also achieve a low KLD when applied naively.

\section{Related Work}
We here focus on approaches that discuss theory-driven analysis of metrics-based evaluation of TG systems that involve human annotations.

\newcite{chaganty-etal-2018-price} propose an approach to combine human and metrics-based evaluation using control variates to reduce biases in automated metrics and save annotation costs. They explore automated scalar metrics in the Summarization and QA domain and find that their approach can lead to marginal reductions of the required human annotations. They conclude that further improvement of automated metrics and exploration of ranking-based evaluation are potential future directions. One interesting take-away from this work is the influence of the quality of the human annotations. Currently, we approximate the oracle preference ratings through human annotations without explicitly modeling uncertainty stemming from annotator disagreement. 
\newcite{Wei2021TheSA} pick up this point and apply a statistical approach to identify a setting in which automated metrics for Machine Translation and Summarization are more precise than human annotations: When the qualitative difference is small and there are only few human annotations. They argue that the reason is that while human annotations are unbiased, they have a high variance. Conversely, automatic metrics tend to have a high bias but low variance. Furthermore, they apply the bias-variance-noise decomposition from \newcite{Domingos2000AUB} to analyse sources of errors in evaluation and asses bias levels in automated metrics. Our analysis, in comparison, is more fine-grained in terms of categorizing metric errors, and we propose how to combine human and metric evaluation under a budget constraint. 
Similar to this work, \newcite{von2022effectiveness} propose a model that captures uncertainties stemming from imperfect metrics and insufficiently sized test sets. With their framework, the required size of a test set that is needed to distinguish a given difference in performance between two systems with a given automated metric can be calculated. Their investigation is limited to the case when scalar metrics are converted to binary metrics, however. 
\newcite{card-etal-2020-little} also analyse the required data set sizes that enable the detection of significant differences between systems, but they do not account for metric errors explicitly.
\newcite{Hashimoto2019UnifyingHA} propose an evaluation approach that combines human and automated evaluation in a statistical framework to estimate diversity and quality of NLG systems jointly. Their focus is the creation of a novel metric,  while our goal is to evaluate existing ones and to combine them with human annotations to obtain robust evaluations.

\section{Conclusion}
In this work, we introduced a novel Bayesian model that explicitly handles errors from automated metrics at the sample level. We then proposed an evaluation protocol that leverages this statistical model to reduce the amount of human annotations needed while yielding similar evaluation outcomes. We applied the protocol to three tasks in a case study. Namely, Dialogue Systems, Summarization, and Machine Translation. The results show that the Bayesian model is able to successfully include various types of uncertainty, which leads to more trustworthy applications of automated metrics. When applying the protocol, we achieve similar results as a purely human evaluation with only half the annotations needed. 

\section*{Acknowledgments}
This work was supported by Swiss National Science Foundation within the project "Virtual Kids - Virtuelle Charaktere zur Verbesserung der Qualität von Kindesbefragungen" [10001A\_189236/1], and internal funding by the Zurich University of Applied Sciences.

\section*{Limitations}
\paragraph{Human Ratings as Oracle.} In this work, we make the strong assumption that human ratings are equivalent to the oracle. As noted in the introduction, human evaluation is hard to setup and does not always lead to satisfactory agreement scores. However, for SummEval and WMT21 the human ratings are provided by experts, and thus, can be seen as close to oracle ratings. For the Dialogue domain, the ratings are made by crowdsourcing where we applied MACE~\cite{hovy2013mace} to get the highest quality ratings. In future work, we will integrate the uncertainty of the human evaluation in the Bayesian model as well, which is not trivial.

\paragraph{Pairwise $\bm{\mu}$.} We noted that the for each pair of systems, the mixture matrix is different. As a consequence, the errors made by the metric must be computed for each pair of systems separately, which is more cost-intensive. In future work, we aim to develop methods to transfer the knowledge from one system pair to another. This also highlights one issue of automated metrics, namely, that they are biased towards certain output types, which are exhibited by certain TG systems. 

\paragraph{Draws are ignored.} One issue with preference-based ratings is the question of how to handle draws on the sample basis. Currently, we use $p_{>}$ and $p_{<}$ to decide if two systems are significantly different. However, if we consider the case of $p_{>}=0.02$, $p_{<}=0.01$, and $p_{=} = 0.97$, then with enough samples, we will measure a statistical significant difference between the two systems. However, in 97\% of cases the outputs are of equal quality. Thus, can we really state that one system is better than the other?

\paragraph{Statistical Significance Decision.} To compare the outcomes of the automated evaluation to the human evaluation, we rely on statistical significance testing. For this, we use the standard approaches, which are widely adopted. However, we noted that the significance decisions are rather arbitrary and make it hard to compare two evaluations, especially the interpretation of Insertion Errors is not trivial. The large amounts of additional automated metric ratings result in some pairs being rated as significantly different. However, it is not clear whether this is a mistake or if we were able to distinguish two systems that were not distinguishable due to too little data. The KLD score gives better insights for this as it compares distributions.

\paragraph{Differences in Samples.} Currently, we disregard the fact that there are samples which are harder to rate than others. In fact, we treat each sample as being equal in Defintion~\ref{def:error_prone_metric}. However, the sample difficulty could be leveraged to distinguish different systems from each other. For instance, if two machine translation systems are evaluated only on easy samples, then they might be rated as being of equal quality. However, a test on a harder sample might show the difference in capabilities between the two systems. 

\paragraph{Conversion to Preference Ratings.} Current automated metrics are built such that they return a scalar value $\in \mathbf{R}$ to rate a given pair of input and output. We have to transform these values into preference ratings by looking at the sign of the difference between the ratings of two outputs (see Appendix~\ref{sec:cs_details}). This leads to a few problems.
First, there are only few draws, since metrics rarely return the exact same floating point value for two different outputs. Second, we disregard the magnitude of the scalar value. The magnitude can be used to assess the certainty of the preference of one output against another. In preliminary experiments we tried including a minimal threshold that the difference needs to surpass in order to be regarded as a preference decision. This will have to be explored in more detail in future work.
%scalar metrics are more powerful than preference metrics: we basically throw out the "magnitude" information of a pairwise comparison which to a certain degree contains a measure of certainty of the metric in its preference

\paragraph{Current Metric performance.} Since the current metrics are not yet of high enough quality, the impact they have on the protocol is small. This might give the impression that the protocol does not offer any remedy. However, the results show that our Bayesian model is able to rectify the overconfidence of low-performance metrics, and in cases where a metric is of low quality, its impact is reduced. 

\bibliography{anthology,custom}

\begin{thebibliography}{57}
\expandafter\ifx\csname natexlab\endcsname\relax\def\natexlab#1{#1}\fi

\bibitem[{Amidei et~al.(2018)Amidei, Piwek, and Willis}]{amidei2018rethinking}
Jacopo Amidei, Paul Piwek, and Alistair Willis. 2018.
\newblock \href {https://aclanthology.org/C18-1281} {Rethinking the agreement
  in human evaluation tasks}.
\newblock In \emph{Proceedings of the 27th International Conference on
  Computational Linguistics}, pages 3318--3329, Santa Fe, New Mexico, USA.
  Association for Computational Linguistics.

\bibitem[{Andrieu et~al.(2003)Andrieu, de~Freitas, Doucet, and
  Jordan}]{intro_mcmc}
Christophe Andrieu, Nando de~Freitas, Arnaud Doucet, and Michael~I. Jordan.
  2003.
\newblock \href {https://doi.org/10.1023/A:1020281327116} {An introduction to
  mcmc for machine learning}.
\newblock 50:5--53.

\bibitem[{Belz et~al.(2021)Belz, Shimorina, Agarwal, and
  Reiter}]{belz2021reprogen}
Anya Belz, Anastasia Shimorina, Shubham Agarwal, and Ehud Reiter. 2021.
\newblock \href {https://aclanthology.org/2021.inlg-1.24} {The {R}epro{G}en
  shared task on reproducibility of human evaluations in {NLG}: Overview and
  results}.
\newblock In \emph{Proceedings of the 14th International Conference on Natural
  Language Generation}, pages 249--258, Aberdeen, Scotland, UK. Association for
  Computational Linguistics.

\bibitem[{Bingham et~al.(2019)Bingham, Chen, Jankowiak, Obermeyer, Pradhan,
  Karaletsos, Singh, Szerlip, Horsfall, and Goodman}]{bingham2019pyro}
Eli Bingham, Jonathan~P. Chen, Martin Jankowiak, Fritz Obermeyer, Neeraj
  Pradhan, Theofanis Karaletsos, Rohit Singh, Paul~A. Szerlip, Paul Horsfall,
  and Noah~D. Goodman. 2019.
\newblock \href {http://jmlr.org/papers/v20/18-403.html} {Pyro: Deep universal
  probabilistic programming}.
\newblock \emph{J. Mach. Learn. Res.}, 20:28:1--28:6.

\bibitem[{Brown et~al.(2020)Brown, Mann, Ryder, Subbiah, Kaplan, Dhariwal,
  Neelakantan, Shyam, Sastry, Askell, Agarwal, Herbert-Voss, Krueger, Henighan,
  Child, Ramesh, Ziegler, Wu, Winter, Hesse, Chen, Sigler, Litwin, Gray, Chess,
  Clark, Berner, McCandlish, Radford, Sutskever, and Amodei}]{brown2020gpt3}
Tom Brown, Benjamin Mann, Nick Ryder, Melanie Subbiah, Jared~D Kaplan, Prafulla
  Dhariwal, Arvind Neelakantan, Pranav Shyam, Girish Sastry, Amanda Askell,
  Sandhini Agarwal, Ariel Herbert-Voss, Gretchen Krueger, Tom Henighan, Rewon
  Child, Aditya Ramesh, Daniel Ziegler, Jeffrey Wu, Clemens Winter, Chris
  Hesse, Mark Chen, Eric Sigler, Mateusz Litwin, Scott Gray, Benjamin Chess,
  Jack Clark, Christopher Berner, Sam McCandlish, Alec Radford, Ilya Sutskever,
  and Dario Amodei. 2020.
\newblock \href
  {https://proceedings.neurips.cc/paper/2020/file/1457c0d6bfcb4967418bfb8ac142f64a-Paper.pdf}
  {Language models are few-shot learners}.
\newblock In \emph{Advances in Neural Information Processing Systems},
  volume~33, pages 1877--1901. Curran Associates, Inc.

\bibitem[{Card et~al.(2020)Card, Henderson, Khandelwal, Jia, Mahowald, and
  Jurafsky}]{card-etal-2020-little}
Dallas Card, Peter Henderson, Urvashi Khandelwal, Robin Jia, Kyle Mahowald, and
  Dan Jurafsky. 2020.
\newblock \href {https://doi.org/10.18653/v1/2020.emnlp-main.745} {With little
  power comes great responsibility}.
\newblock In \emph{Proceedings of the 2020 Conference on Empirical Methods in
  Natural Language Processing (EMNLP)}, pages 9263--9274, Online. Association
  for Computational Linguistics.

\bibitem[{Celikyilmaz et~al.(2020)Celikyilmaz, Clark, and
  Gao}]{celikyilmaz2020evaluation}
Asli Celikyilmaz, Elizabeth Clark, and Jianfeng Gao. 2020.
\newblock Evaluation of text generation: A survey.
\newblock \emph{arXiv preprint arXiv:2006.14799}.

\bibitem[{Chaganty et~al.(2018)Chaganty, Mussmann, and
  Liang}]{chaganty-etal-2018-price}
Arun Chaganty, Stephen Mussmann, and Percy Liang. 2018.
\newblock \href {https://doi.org/10.18653/v1/P18-1060} {The price of debiasing
  automatic metrics in natural language evalaution}.
\newblock In \emph{Proceedings of the 56th Annual Meeting of the Association
  for Computational Linguistics (Volume 1: Long Papers)}, pages 643--653,
  Melbourne, Australia. Association for Computational Linguistics.

\bibitem[{Coakley and Heise(1996)}]{sign_tests}
Clint~W. Coakley and Mark~A. Heise. 1996.
\newblock \href {http://www.jstor.org/stable/2532840} {Versions of the sign
  test in the presence of ties}.
\newblock \emph{Biometrics}, 52(4):1242--1251.

\bibitem[{Deriu et~al.(2021)Deriu, Rodrigo, Otegi, Echegoyen, Rosset, Agirre,
  and Cieliebak}]{deriu2021survey}
Jan Deriu, Alvaro Rodrigo, Arantxa Otegi, Guillermo Echegoyen, Sophie Rosset,
  Eneko Agirre, and Mark Cieliebak. 2021.
\newblock Survey on evaluation methods for dialogue systems.
\newblock \emph{Artificial Intelligence Review}, 54(1):755--810.

\bibitem[{Devlin et~al.(2019)Devlin, Chang, Lee, and
  Toutanova}]{devlin2019bert}
Jacob Devlin, Ming-Wei Chang, Kenton Lee, and Kristina Toutanova. 2019.
\newblock \href {https://doi.org/10.18653/v1/N19-1423} {{BERT}: Pre-training of
  deep bidirectional transformers for language understanding}.
\newblock In \emph{Proceedings of the 2019 Conference of the North {A}merican
  Chapter of the Association for Computational Linguistics: Human Language
  Technologies, Volume 1 (Long and Short Papers)}, pages 4171--4186,
  Minneapolis, Minnesota. Association for Computational Linguistics.

\bibitem[{Domingos(2000)}]{Domingos2000AUB}
Pedro~M. Domingos. 2000.
\newblock A unified bias-variance decomposition and its applications.

\bibitem[{Fabbri et~al.(2021)Fabbri, Kry{\'s}ci{\'n}ski, McCann, Xiong, Socher,
  and Radev}]{fabbri2021summeval}
Alexander~R. Fabbri, Wojciech Kry{\'s}ci{\'n}ski, Bryan McCann, Caiming Xiong,
  Richard Socher, and Dragomir Radev. 2021.
\newblock \href {https://doi.org/10.1162/tacl_a_00373} {{S}umm{E}val:
  Re-evaluating summarization evaluation}.
\newblock \emph{Transactions of the Association for Computational Linguistics},
  9:391--409.

\bibitem[{Freitag et~al.(2021)Freitag, Rei, Mathur, Lo, Stewart, Foster, Lavie,
  and Bojar}]{freitag2021wmt21metrics}
Markus Freitag, Ricardo Rei, Nitika Mathur, Chi-kiu Lo, Craig Stewart, George
  Foster, Alon Lavie, and Ond{\v{r}}ej Bojar. 2021.
\newblock \href {https://aclanthology.org/2021.wmt-1.73} {Results of the
  {WMT}21 metrics shared task: Evaluating metrics with expert-based human
  evaluations on {TED} and news domain}.
\newblock In \emph{Proceedings of the Sixth Conference on Machine Translation},
  pages 733--774, Online. Association for Computational Linguistics.

\bibitem[{Gao et~al.(2020)Gao, Zhao, and Eger}]{gao2020supert}
Yang Gao, Wei Zhao, and Steffen Eger. 2020.
\newblock \href {https://doi.org/10.18653/v1/2020.acl-main.124} {{SUPERT}:
  Towards new frontiers in unsupervised evaluation metrics for multi-document
  summarization}.
\newblock In \emph{Proceedings of the 58th Annual Meeting of the Association
  for Computational Linguistics}, pages 1347--1354, Online. Association for
  Computational Linguistics.

\bibitem[{Hashimoto et~al.(2019)Hashimoto, Zhang, and
  Liang}]{Hashimoto2019UnifyingHA}
Tatsunori~B. Hashimoto, Hugh Zhang, and Percy Liang. 2019.
\newblock Unifying human and statistical evaluation for natural language
  generation.
\newblock In \emph{North American Chapter of the Association for Computational
  Linguistics}.

\bibitem[{Hoffman and Gelman(2014)}]{JMLR:v15:hoffman14a}
Matthew~D. Hoffman and Andrew Gelman. 2014.
\newblock \href {http://jmlr.org/papers/v15/hoffman14a.html} {The no-u-turn
  sampler: Adaptively setting path lengths in hamiltonian monte carlo}.
\newblock \emph{Journal of Machine Learning Research}, 15(47):1593--1623.

\bibitem[{Hovy et~al.(2013)Hovy, Berg-Kirkpatrick, Vaswani, and
  Hovy}]{hovy2013mace}
Dirk Hovy, Taylor Berg-Kirkpatrick, Ashish Vaswani, and Eduard Hovy. 2013.
\newblock \href {https://aclanthology.org/N13-1132} {Learning whom to trust
  with {MACE}}.
\newblock In \emph{Proceedings of the 2013 Conference of the North {A}merican
  Chapter of the Association for Computational Linguistics: Human Language
  Technologies}, pages 1120--1130, Atlanta, Georgia. Association for
  Computational Linguistics.

\bibitem[{Huang et~al.(2020)Huang, Ye, Qin, Lin, and Liang}]{huang2020-grade}
Lishan Huang, Zheng Ye, Jinghui Qin, Liang Lin, and Xiaodan Liang. 2020.
\newblock \href {https://doi.org/10.18653/v1/2020.emnlp-main.742} {{GRADE}:
  Automatic graph-enhanced coherence metric for evaluating open-domain dialogue
  systems}.
\newblock In \emph{Proceedings of the 2020 Conference on Empirical Methods in
  Natural Language Processing (EMNLP)}, pages 9230--9240, Online. Association
  for Computational Linguistics.

\bibitem[{Humeau et~al.(2020)Humeau, Shuster, Lachaux, and
  Weston}]{Humeau2020Poly-encoders}
Samuel Humeau, Kurt Shuster, Marie-Anne Lachaux, and Jason Weston. 2020.
\newblock \href {https://openreview.net/forum?id=SkxgnnNFvH} {Poly-encoders:
  Architectures and pre-training strategies for fast and accurate
  multi-sentence scoring}.
\newblock In \emph{International Conference on Learning Representations}.

\bibitem[{Kocmi et~al.(2021)Kocmi, Federmann, Grundkiewicz, Junczys-Dowmunt,
  Matsushita, and Menezes}]{kocmi-etal-2021-ship}
Tom Kocmi, Christian Federmann, Roman Grundkiewicz, Marcin Junczys-Dowmunt,
  Hitokazu Matsushita, and Arul Menezes. 2021.
\newblock \href {https://aclanthology.org/2021.wmt-1.57} {To ship or not to
  ship: An extensive evaluation of automatic metrics for machine translation}.
\newblock In \emph{Proceedings of the Sixth Conference on Machine Translation},
  pages 478--494, Online. Association for Computational Linguistics.

\bibitem[{Komeili et~al.(2022)Komeili, Shuster, and
  Weston}]{komeili2022BB2internet}
Mojtaba Komeili, Kurt Shuster, and Jason Weston. 2022.
\newblock \href {https://doi.org/10.18653/v1/2022.acl-long.579}
  {{I}nternet-augmented dialogue generation}.
\newblock In \emph{Proceedings of the 60th Annual Meeting of the Association
  for Computational Linguistics (Volume 1: Long Papers)}, pages 8460--8478,
  Dublin, Ireland. Association for Computational Linguistics.

\bibitem[{Lin(2004)}]{lin2004rouge}
Chin-Yew Lin. 2004.
\newblock \href {http://www.aclweb.org/anthology/W04-1013} {{ROUGE: A Package
  for Automatic Evaluation of Summaries}}.
\newblock In \emph{Text Summarization Branches Out: Proceedings of the ACL-04
  Workshop}, pages 74--81, Barcelona, Spain. Association for Computational
  Linguistics.

\bibitem[{Lommel et~al.(2014)Lommel, Uszkoreit, and Burchardt}]{lommel2014mqm}
Arle Lommel, Hans Uszkoreit, and Aljoscha Burchardt. 2014.
\newblock Multidimensional quality metrics (mqm): A framework for declaring and
  describing translation quality metrics.
\newblock \emph{Revista Tradum{\`a}tica: tecnologies de la traducci{\'o}},
  (12):455--463.

\bibitem[{Lowe et~al.(2017)Lowe, Noseworthy, Serban, Angelard-Gontier, Bengio,
  and Pineau}]{lowe2017adem}
Ryan Lowe, Michael Noseworthy, Iulian~Vlad Serban, Nicolas Angelard-Gontier,
  Yoshua Bengio, and Joelle Pineau. 2017.
\newblock \href {https://doi.org/10.18653/v1/P17-1103} {Towards an automatic
  {T}uring test: Learning to evaluate dialogue responses}.
\newblock In \emph{Proceedings of the 55th Annual Meeting of the Association
  for Computational Linguistics (Volume 1: Long Papers)}, pages 1116--1126,
  Vancouver, Canada. Association for Computational Linguistics.

\bibitem[{Mathur et~al.(2020)Mathur, Baldwin, and
  Cohn}]{mathur-etal-2020-tangled}
Nitika Mathur, Timothy Baldwin, and Trevor Cohn. 2020.
\newblock \href {https://doi.org/10.18653/v1/2020.acl-main.448} {Tangled up in
  {BLEU}: Reevaluating the evaluation of automatic machine translation
  evaluation metrics}.
\newblock In \emph{Proceedings of the 58th Annual Meeting of the Association
  for Computational Linguistics}, pages 4984--4997, Online. Association for
  Computational Linguistics.

\bibitem[{Mehri and Eskenazi(2020{\natexlab{a}})}]{mehri2020fed}
Shikib Mehri and Maxine Eskenazi. 2020{\natexlab{a}}.
\newblock \href {https://aclanthology.org/2020.sigdial-1.28} {Unsupervised
  evaluation of interactive dialog with {D}ialo{GPT}}.
\newblock In \emph{Proceedings of the 21th Annual Meeting of the Special
  Interest Group on Discourse and Dialogue}, pages 225--235, 1st virtual
  meeting. Association for Computational Linguistics.

\bibitem[{Mehri and Eskenazi(2020{\natexlab{b}})}]{mehri2020usr}
Shikib Mehri and Maxine Eskenazi. 2020{\natexlab{b}}.
\newblock \href {https://doi.org/10.18653/v1/2020.acl-main.64} {{USR: An
  Unsupervised and Reference Free Evaluation Metric for Dialog Generation}}.
\newblock In \emph{Proceedings of the 58th Annual Meeting of the Association
  for Computational Linguistics}, pages 681--707, Online. Association for
  Computational Linguistics.

\bibitem[{Metropolis and Ulam(1949)}]{monte_carlo}
Nicholas Metropolis and S.~Ulam. 1949.
\newblock \href {http://www.jstor.org/stable/2280232} {The monte carlo method}.
\newblock \emph{Journal of the American Statistical Association},
  44(247):335--341.

\bibitem[{{Miller} et~al.(2017){Miller}, {Feng}, {Fisch}, {Lu}, {Batra},
  {Bordes}, {Parikh}, and {Weston}}]{miller2017parlai}
A.~H. {Miller}, W.~{Feng}, A.~{Fisch}, J.~{Lu}, D.~{Batra}, A.~{Bordes},
  D.~{Parikh}, and J.~{Weston}. 2017.
\newblock Parlai: A dialog research software platform.
\newblock \emph{arXiv preprint arXiv:{1705.06476}}.

\bibitem[{Nallapati et~al.(2016)Nallapati, Zhou, dos Santos, Gul{\c{c}}ehre,
  and Xiang}]{nallapati2016abstractive}
Ramesh Nallapati, Bowen Zhou, Cicero dos Santos, {\c{C}}a{\u{g}}lar
  Gul{\c{c}}ehre, and Bing Xiang. 2016.
\newblock \href {https://doi.org/10.18653/v1/K16-1028} {Abstractive text
  summarization using sequence-to-sequence {RNN}s and beyond}.
\newblock In \emph{Proceedings of the 20th {SIGNLL} Conference on Computational
  Natural Language Learning}, pages 280--290, Berlin, Germany. Association for
  Computational Linguistics.

\bibitem[{Pang et~al.(2020)Pang, Nijkamp, Han, Zhou, Liu, and
  Tu}]{pang2020towards}
Bo~Pang, Erik Nijkamp, Wenjuan Han, Linqi Zhou, Yixian Liu, and Kewei Tu. 2020.
\newblock \href {https://doi.org/10.18653/v1/2020.acl-main.333} {Towards
  holistic and automatic evaluation of open-domain dialogue generation}.
\newblock In \emph{Proceedings of the 58th Annual Meeting of the Association
  for Computational Linguistics}, pages 3619--3629, Online. Association for
  Computational Linguistics.

\bibitem[{Papineni et~al.(2002)Papineni, Roukos, Ward, and
  Zhu}]{papineni2002bleu}
Kishore Papineni, Salim Roukos, Todd Ward, and Wei-Jing Zhu. 2002.
\newblock \href {https://doi.org/10.3115/1073083.1073135} {{Bleu: a Method for
  Automatic Evaluation of Machine Translation}}.
\newblock In \emph{Proceedings of the 40th Annual Meeting of the Association
  for Computational Linguistics}, pages 311--318, Philadelphia, Pennsylvania,
  USA. Association for Computational Linguistics.

\bibitem[{Peyrard et~al.(2017)Peyrard, Botschen, and Gurevych}]{peyrard2017s3}
Maxime Peyrard, Teresa Botschen, and Iryna Gurevych. 2017.
\newblock \href {https://doi.org/10.18653/v1/W17-4510} {Learning to score
  system summaries for better content selection evaluation.}
\newblock In \emph{Proceedings of the Workshop on New Frontiers in
  Summarization}, pages 74--84, Copenhagen, Denmark. Association for
  Computational Linguistics.

\bibitem[{Phan et~al.(2019)Phan, Pradhan, and Jankowiak}]{numpyro1}
Du~Phan, Neeraj Pradhan, and Martin Jankowiak. 2019.
\newblock Composable effects for flexible and accelerated probabilistic
  programming in numpyro.
\newblock \emph{arXiv preprint arXiv:1912.11554}.

\bibitem[{Phy et~al.(2020)Phy, Zhao, and Aizawa}]{phy2020deconstruct}
Vitou Phy, Yang Zhao, and Akiko Aizawa. 2020.
\newblock \href {https://doi.org/10.18653/v1/2020.coling-main.368} {Deconstruct
  to reconstruct a configurable evaluation metric for open-domain dialogue
  systems}.
\newblock In \emph{Proceedings of the 28th International Conference on
  Computational Linguistics}, pages 4164--4178, Barcelona, Spain (Online).
  International Committee on Computational Linguistics.

\bibitem[{Radford et~al.()Radford, Wu, Child, Luan, Amodei, Sutskever
  et~al.}]{radford2019gpt2}
Alec Radford, Jeffrey Wu, Rewon Child, David Luan, Dario Amodei, Ilya
  Sutskever, et~al.
\newblock Language models are unsupervised multitask learners.

\bibitem[{Raffel et~al.(2020)Raffel, Shazeer, Roberts, Lee, Narang, Matena,
  Zhou, Li, and Liu}]{reffel2020t5}
Colin Raffel, Noam Shazeer, Adam Roberts, Katherine Lee, Sharan Narang, Michael
  Matena, Yanqi Zhou, Wei Li, and Peter~J. Liu. 2020.
\newblock \href {http://jmlr.org/papers/v21/20-074.html} {Exploring the limits
  of transfer learning with a unified text-to-text transformer}.
\newblock \emph{Journal of Machine Learning Research}, 21(140):1--67.

\bibitem[{Rei et~al.(2020)Rei, Stewart, Farinha, and Lavie}]{rei2020comet}
Ricardo Rei, Craig Stewart, Ana~C Farinha, and Alon Lavie. 2020.
\newblock \href {https://doi.org/10.18653/v1/2020.emnlp-main.213} {{COMET}: A
  neural framework for {MT} evaluation}.
\newblock In \emph{Proceedings of the 2020 Conference on Empirical Methods in
  Natural Language Processing (EMNLP)}, pages 2685--2702, Online. Association
  for Computational Linguistics.

\bibitem[{Sai et~al.(2020)Sai, Mohankumar, Arora, and Khapra}]{sai2020deb}
Ananya~B. Sai, Akash~Kumar Mohankumar, Siddhartha Arora, and Mitesh~M. Khapra.
  2020.
\newblock \href {https://doi.org/10.1162/tacl_a_00347} {{Improving Dialog
  Evaluation with a Multi-reference Adversarial Dataset and Large Scale
  Pretraining}}.
\newblock \emph{Transactions of the Association for Computational Linguistics},
  8:810--827.

\bibitem[{Scialom et~al.(2019)Scialom, Lamprier, Piwowarski, and
  Staiano}]{scialom2019summaqa}
Thomas Scialom, Sylvain Lamprier, Benjamin Piwowarski, and Jacopo Staiano.
  2019.
\newblock \href {https://doi.org/10.18653/v1/D19-1320} {Answers unite!
  unsupervised metrics for reinforced summarization models}.
\newblock In \emph{Proceedings of the 2019 Conference on Empirical Methods in
  Natural Language Processing and the 9th International Joint Conference on
  Natural Language Processing (EMNLP-IJCNLP)}, pages 3246--3256, Hong Kong,
  China. Association for Computational Linguistics.

\bibitem[{Sellam et~al.(2020)Sellam, Das, and Parikh}]{sellam2020bleurt}
Thibault Sellam, Dipanjan Das, and Ankur Parikh. 2020.
\newblock \href {https://doi.org/10.18653/v1/2020.acl-main.704} {{BLEURT}:
  Learning robust metrics for text generation}.
\newblock In \emph{Proceedings of the 58th Annual Meeting of the Association
  for Computational Linguistics}, pages 7881--7892, Online. Association for
  Computational Linguistics.

\bibitem[{Shuster et~al.(2022{\natexlab{a}})Shuster, Komeili, Adolphs, Roller,
  Szlam, and Weston}]{shuster2022seeker}
Kurt Shuster, Mojtaba Komeili, Leonard Adolphs, Stephen Roller, Arthur Szlam,
  and Jason Weston. 2022{\natexlab{a}}.
\newblock Language models that seek for knowledge: Modular search \& generation
  for dialogue and prompt completion.
\newblock \emph{arXiv preprint arXiv:2203.13224}.

\bibitem[{Shuster et~al.(2022{\natexlab{b}})Shuster, Xu, Komeili, Ju, Smith,
  Roller, Ung, Chen, Arora, Lane et~al.}]{shuster2022blenderbot}
Kurt Shuster, Jing Xu, Mojtaba Komeili, Da~Ju, Eric~Michael Smith, Stephen
  Roller, Megan Ung, Moya Chen, Kushal Arora, Joshua Lane, et~al.
  2022{\natexlab{b}}.
\newblock Blenderbot 3: a deployed conversational agent that continually learns
  to responsibly engage.
\newblock \emph{arXiv preprint arXiv:2208.03188}.

\bibitem[{Sinha et~al.(2020)Sinha, Parthasarathi, Wang, Lowe, Hamilton, and
  Pineau}]{sinha2020maude}
Koustuv Sinha, Prasanna Parthasarathi, Jasmine Wang, Ryan Lowe, William~L.
  Hamilton, and Joelle Pineau. 2020.
\newblock \href {https://doi.org/10.18653/v1/2020.acl-main.220} {Learning an
  unreferenced metric for online dialogue evaluation}.
\newblock In \emph{Proceedings of the 58th Annual Meeting of the Association
  for Computational Linguistics}, pages 2430--2441, Online. Association for
  Computational Linguistics.

\bibitem[{Smith et~al.(2020)Smith, Williamson, Shuster, Weston, and
  Boureau}]{smith-etal-2020-BST}
Eric~Michael Smith, Mary Williamson, Kurt Shuster, Jason Weston, and Y-Lan
  Boureau. 2020.
\newblock \href {https://doi.org/10.18653/v1/2020.acl-main.183} {Can you put it
  all together: Evaluating conversational agents{'} ability to blend skills}.
\newblock In \emph{Proceedings of the 58th Annual Meeting of the Association
  for Computational Linguistics}, pages 2021--2030, Online. Association for
  Computational Linguistics.

\bibitem[{Takahashi et~al.(2021)Takahashi, Ishibashi, Sudoh, and
  Nakamura}]{takahashi2021cspec}
Kosuke Takahashi, Yoichi Ishibashi, Katsuhito Sudoh, and Satoshi Nakamura.
  2021.
\newblock \href {https://aclanthology.org/2021.wmt-1.113} {Multilingual machine
  translation evaluation metrics fine-tuned on pseudo-negative examples for
  {WMT} 2021 metrics task}.
\newblock In \emph{Proceedings of the Sixth Conference on Machine Translation},
  pages 1049--1052, Online. Association for Computational Linguistics.

\bibitem[{Vasilyev et~al.(2020)Vasilyev, Dharnidharka, and
  Bohannon}]{vasilyev2020blanc}
Oleg Vasilyev, Vedant Dharnidharka, and John Bohannon. 2020.
\newblock \href {https://doi.org/10.18653/v1/2020.eval4nlp-1.2} {Fill in the
  {BLANC}: Human-free quality estimation of document summaries}.
\newblock In \emph{Proceedings of the First Workshop on Evaluation and
  Comparison of NLP Systems}, pages 11--20, Online. Association for
  Computational Linguistics.

\bibitem[{Vaswani et~al.(2017)Vaswani, Shazeer, Parmar, Uszkoreit, Jones,
  Gomez, Kaiser, and Polosukhin}]{vaswani2017attention}
Ashish Vaswani, Noam Shazeer, Niki Parmar, Jakob Uszkoreit, Llion Jones,
  Aidan~N Gomez, Lukasz Kaiser, and Illia Polosukhin. 2017.
\newblock \href
  {https://proceedings.neurips.cc/paper/2017/file/3f5ee243547dee91fbd053c1c4a845aa-Paper.pdf}
  {Attention is all you need}.
\newblock In \emph{Advances in Neural Information Processing Systems},
  volume~30. Curran Associates, Inc.

\bibitem[{Vedantam et~al.(2015)Vedantam, Lawrence~Zitnick, and
  Parikh}]{Vedantam2015cider}
Ramakrishna Vedantam, C.~Lawrence~Zitnick, and Devi Parikh. 2015.
\newblock Cider: Consensus-based image description evaluation.
\newblock In \emph{Proceedings of the IEEE Conference on Computer Vision and
  Pattern Recognition (CVPR)}.

\bibitem[{von D{\"a}niken et~al.(2022)von D{\"a}niken, Deriu, Tuggener, and
  Cieliebak}]{von2022effectiveness}
Pius von D{\"a}niken, Jan Deriu, Don Tuggener, and Mark Cieliebak. 2022.
\newblock On the effectiveness of automated metrics for text generation
  systems.
\newblock \emph{arXiv preprint arXiv:2210.13025}.

\bibitem[{Walker et~al.(1997)Walker, Litman, Kamm, and
  Abella}]{walker1997paradise}
Marilyn~A. Walker, Diane~J. Litman, Candace~A. Kamm, and Alicia Abella. 1997.
\newblock \href {https://doi.org/10.3115/976909.979652} {{PARADISE}: A
  framework for evaluating spoken dialogue agents}.
\newblock In \emph{35th Annual Meeting of the Association for Computational
  Linguistics and 8th Conference of the {E}uropean Chapter of the Association
  for Computational Linguistics}, pages 271--280, Madrid, Spain. Association
  for Computational Linguistics.

\bibitem[{Wei and Jia(2021)}]{Wei2021TheSA}
Johnny Tian-Zheng Wei and Robin Jia. 2021.
\newblock The statistical advantage of automatic nlg metrics at the system
  level.
\newblock In \emph{Annual Meeting of the Association for Computational
  Linguistics}.

\bibitem[{Xu et~al.(2022)Xu, Szlam, and Weston}]{xu2022BB2beyond}
Jing Xu, Arthur Szlam, and Jason Weston. 2022.
\newblock \href {https://doi.org/10.18653/v1/2022.acl-long.356} {Beyond
  goldfish memory: Long-term open-domain conversation}.
\newblock In \emph{Proceedings of the 60th Annual Meeting of the Association
  for Computational Linguistics (Volume 1: Long Papers)}, pages 5180--5197,
  Dublin, Ireland. Association for Computational Linguistics.

\bibitem[{Yeh et~al.(2021)Yeh, Eskenazi, and Mehri}]{yeh2021comprehensive}
Yi-Ting Yeh, Maxine Eskenazi, and Shikib Mehri. 2021.
\newblock \href {https://doi.org/10.18653/v1/2021.eancs-1.3} {A comprehensive
  assessment of dialog evaluation metrics}.
\newblock In \emph{The First Workshop on Evaluations and Assessments of Neural
  Conversation Systems}, pages 15--33, Online. Association for Computational
  Linguistics.

\bibitem[{Zhang et~al.(2019)Zhang, Kishore, Wu, Weinberger, and
  Artzi}]{zhang2019bertscore}
Tianyi Zhang, Varsha Kishore, Felix Wu, Kilian~Q Weinberger, and Yoav Artzi.
  2019.
\newblock Bertscore: Evaluating text generation with bert.
\newblock \emph{arXiv preprint arXiv:1904.09675}.

\bibitem[{Zhang et~al.(2020)Zhang, Sun, Galley, Chen, Brockett, Gao, Gao, Liu,
  and Dolan}]{zhang2020dialogpt}
Yizhe Zhang, Siqi Sun, Michel Galley, Yen-Chun Chen, Chris Brockett, Xiang Gao,
  Jianfeng Gao, Jingjing Liu, and Bill Dolan. 2020.
\newblock \href {https://doi.org/10.18653/v1/2020.acl-demos.30} {{DIALOGPT} :
  Large-scale generative pre-training for conversational response generation}.
\newblock In \emph{Proceedings of the 58th Annual Meeting of the Association
  for Computational Linguistics: System Demonstrations}, pages 270--278,
  Online. Association for Computational Linguistics.

\end{thebibliography}
\bibliographystyle{acl_natbib}

\appendix

\section{Derivations}
\label{sec:derivations}

\paragraph{Dirichlet.} We will first explain the usage of Dirichlet distributions in Equations~\ref{eq:oracle_posterior} and~\ref{eq:mu:diriclet}. The Dirichlet distribution of order $K$ is defined for all $K$ dimensional probability vectors $\bm{p} = (p_1, \dots, p_K)$ such that $\sum_{i=1}^{K}p_i = 1$ and $p_i \ge 0$. It has $K$ parameters $\bm{\alpha} = (\alpha_1, \dots, \alpha_K)$ and its density is:
\[
p(\bm{p} | \bm{\alpha}) = \frac{1}{B(\bm{\alpha})}\prod_{i=1}^{K}p_{i}^{\alpha_i - 1}
\],
where $B(\bm{\alpha})$ is the multivariate Beta function used to normalize the distribution. Note that if all $\alpha_i = 1$ then the distribution is constant at all points $\bm{p}$, meaning it is equivalent to the uniform distribution in that case.
Our main interest in using the Dirichlet distribution is because it is the conjugate prior of the Multinomial distribution. In Section~\ref{sec:framework} the counts from the oracle ratings $n_>$, $n_=$, and $n_<$ follow a Multinomial distribution with unknown probabilities $\bm{p} = (p_>, p_=, p_<)$, meaning that:
\begin{equation*}
{\small
    \begin{aligned}
        &P(N_+ = n_+, N_= = n_=, N_< = n_< | \bm{p}) \\
        &= \frac{(n_> + n_= + n_<)}{n_>!n_=!n_<!}p_{>}^{n_>}p_{=}^{n_=}p_{<}^{n_<}
    \end{aligned}
    }
\end{equation*}
If we assume a Dirichlet prior for $\bm{p} \sim Dirichlet(\alpha_>, \alpha_=, \alpha_<)$ then we can compute its posterior:
\begin{equation*}
{\small
\begin{aligned}
    &p(\bm{p} | N) \propto P(N | \bm{p})p(\bm{p}) \\
    &\propto p_{>}^{n_>}p_{=}^{n_=}p_{<}^{n_<} p_{>}^{\alpha_> -1 }p_{=}^{\alpha_= -1}p_{<}^{\alpha_< - 1} \\
    &\propto p_{>}^{\alpha_< + n_< - 1}p_{=}^{\alpha_= + n_= -1}p_{<}^{\alpha_< + n_< -1}
\end{aligned}
}
\end{equation*}
We left out the normalization constants in this derivation. We can see that on the final line we arrive at the density of an updated Dirichlet distribution $Dirichlet(\alpha_> + n_>, \alpha_= + n_=, \alpha_< + n_<)$. Setting $\alpha_i = 1$ we get our result in Equation~\ref{eq:oracle_posterior}.

We can apply the same principle to the columns of $\bm{\mu}$. The first column $\bm{\mu}_{\cdot>} = (\mu_{>>}, \mu_{=>}, \mu_{<>})^{T}$ denotes the conditional probabilities of getting a specific outcome from the metric conditioned on the oracle rating being $>$. The associated confusion counts $n_{>>}$, $n_{=>}$, and $n_{<>}$ again follow a Trinomial distribution with outcome probabilities of $\bm{\mu}_{\cdot>}$. If we again assume a uniform prior for $\bm{\mu}_{\cdot>}$ then we can derive the posteriors in Equation~\ref{eq:mu:diriclet}.

\paragraph{Mixture.} In Sections~\ref{sec:step_two} and~\ref{sec:step_three} we use the fact that the probabilities associated with the counts of metric ratings is $\hat{\bm{p}} = \bm{\mu}\bm{p}$. We know that $\hat{p}_> = P(\mathcal{M}(i_j, o_{j}^{a}, o_{j}^{b}) = \text{>})$. Using the law of total probability:
\begin{equation*}
{\small
\begin{aligned}
\hat{p}_> = \sum_{c \in \{>,=,<\}} & P(\mathcal{M}(i_j, o_{j}^{a}, o_{j}^{b}) = \text{>} | \Omega(i_j, o_{j}^{a}, o_{j}^{b}) = c) \\
& P(\Omega(i_j, o_{j}^{a}, o_{j}^{b}) = c)
\end{aligned}
}
\end{equation*}
We note that $P(\Omega(i_j, o_{j}^{a}, o_{j}^{b}) = c) = p_c$, and therefore $\hat{p}_> = \sum_{c \in \{>,=,<\}}\mu_{>c}p_c$ and analogously for $\hat{p}_=$ and $\hat{p}_<$. This leads us to our original statement $\hat{\bm{p}} = \bm{\mu}\bm{p}$.

\paragraph{Using Annotations multiple times.} In Algorithm~\ref{algo:decision} it can be unclear which subsets of annotations should be used for the various counts. In general, the test set of inputs $\mathcal{I}$ can be split into three subsets: $\mathcal{I}_{M, A}$, the samples for which we have paired ratings from both the metric and humans, $\mathcal{I}_{M}$, the samples for which we have only ratings from the automated metric, and $\mathcal{I}_{A}$, the samples for which we only have human ratings. It is relatively obvious that we can use $\mathcal{I}_{M}$ to count $m_>, m_=, m_<$, $\mathcal{I}_{A}$ for $n_>, n_=, n_<$, and $\mathcal{I}_{M, A}$ for the confusion counts $n_{cc'}$. The question is whether it is sound to use the ratings from $\mathcal{I}_{M, A}$ to augment $n_c$ and $m_c$. In general, it should not be an issue to use the human ratings in $\mathcal{I}_{M, A}$ as additional counts for $n_>, n_=, n_<$. But we must not use the metric ratings to get additional counts $m_>, m_=, m_<$. This means that, in principal, $\mathcal{I}_{A}$ could be empty.

\section{Markov Chain Monte Carlo (MCMC) Sampling}
\label{sec:mcmc_samp}
Markov Chain Monte Carlo (MCMC) methods are often used in Bayesian modelling when there is no analytic closed form solution for the resulting posterior. The main idea is that expected values of functions of the posterior can be reasonably approximated by averaging over samples drawn from it~\citep{monte_carlo}. Samples are generated sequentially, and the next sample is usually generated by first modifying the current sample randomly and then either accepting or rejecting it based on its likelihood. We refer the interested reader to~\citet{intro_mcmc} for an introduction.

We use the \textit{Numpyro}~\citep{bingham2019pyro, numpyro1} library to implement the framework laid out in Section~\ref{sec:framework}. We use the built-in No-U-Turn (NUTS) Sampler~\citep{JMLR:v15:hoffman14a}. When running the decision function laid out in Algorithm~\ref{algo:decision}, we run 5 chains in parallel. We use a warm-up period of 2000 samples per chain, which are discarded, and draw 10000 samples per chain to keep and compute the difference in win rates.

\section{Case Study Details}
\label{sec:cs_details}
For the case studies, we require two types of ratings: preference ratings made by humans to simulate the oracle $\Omega$, and metric ratings for the error-prone ratings $\mathcal{M}$. We collect this data for three types of domains: Conversational Dialogue Systems, Automated Text Summarization, and Machine Translation. 

Since most metrics return a scalar value, we need to transform them into a preference rating, which is done as follows:

\begin{definition}[Scalar Metric]
    We call real valued functions of inputs and outputs \textbf{scalar metrics}:
    $\mathcal{M}_s: \mathcal{I} \times \mathcal{O} \rightarrow \mathbb{R}$.
\end{definition}

A preference metric can be constructed from a scalar metric as follows:
\begin{definition}
\label{def:derived_metric}
    The \textbf{derived comparison metric} $\mathcal{M}$ of a given scalar metric $\mathcal{M}_{s}$
    is defined as
    \small
    \[
        \mathcal{M}(i, o_1, o_2) =
        \begin{cases}
            > & \mathcal{M}_{s}(i, o_1) > \mathcal{M}_{s}(i, o_2) \\
            = & \textit{otherwise} \\
            < & \mathcal{M}_{s}(i, o_1) < \mathcal{M}_{s}(i, o_2) \\
        \end{cases}
    \]
\end{definition}

\subsection{Dialog System Data Collection}
For the Dialog domain, we used the ParlAI framework~\cite{miller2017parlai} to generate the outputs of systems. For this, we selected 5 state-of-the-art dialog systems, and used ParlAI to generate response for a static context. The Dialogue Systems are:
\begin{itemize}[noitemsep,topsep=0pt]
    \item \textbf{Blenderbot 1.0 - 400distill (BL400distill)}. BlenderBot 1.0~\cite{shuster2022blenderbot} is a transformer-based encoder-decoder bot trained on the Blended Skill Task (BST) data~\cite{smith-etal-2020-BST}. 
    \item \textbf{Blenderbot 2.0 - 3B Params (BL2-3B)}. The BlenderBot 2.0 extends Blenderbot 1.0 with internet access~\cite{komeili2022BB2internet} and a long-term memory~\cite{xu2022BB2beyond}. 
    \item \textbf{DialoGPT}. DialoGPT~\cite{zhang2020dialogpt} is a decoder-only dialogue system that fine-tunes GPT-2~\cite{radford2019gpt2} on the Reddit dataset proposed by the DialoGPT authors.
    \item \textbf{PolyEncoder}. The PolyEnocder~\cite{Humeau2020Poly-encoders} is a retrieval based dialogue system, which selects the most suitable response from a set of candidates. 
    \item \textbf{SeekerDial3B}. SeekerDial~\cite{shuster2022seeker} improves on the internet search proposed in BlenderBot 2.0
\end{itemize}

We used ParlAI to generate the response of each of the dialogue systems for 1000 static contexts from the Blended Skill Task (BST) test set. From this set, we selected 50 contexts, and generated all pairwise outputs between all 5 dialogue systems and the human reference. That is, for each context, there are 15 pairs of outputs to be rated. 

\begin{figure*}
    \centering
    \includegraphics[width=\textwidth]{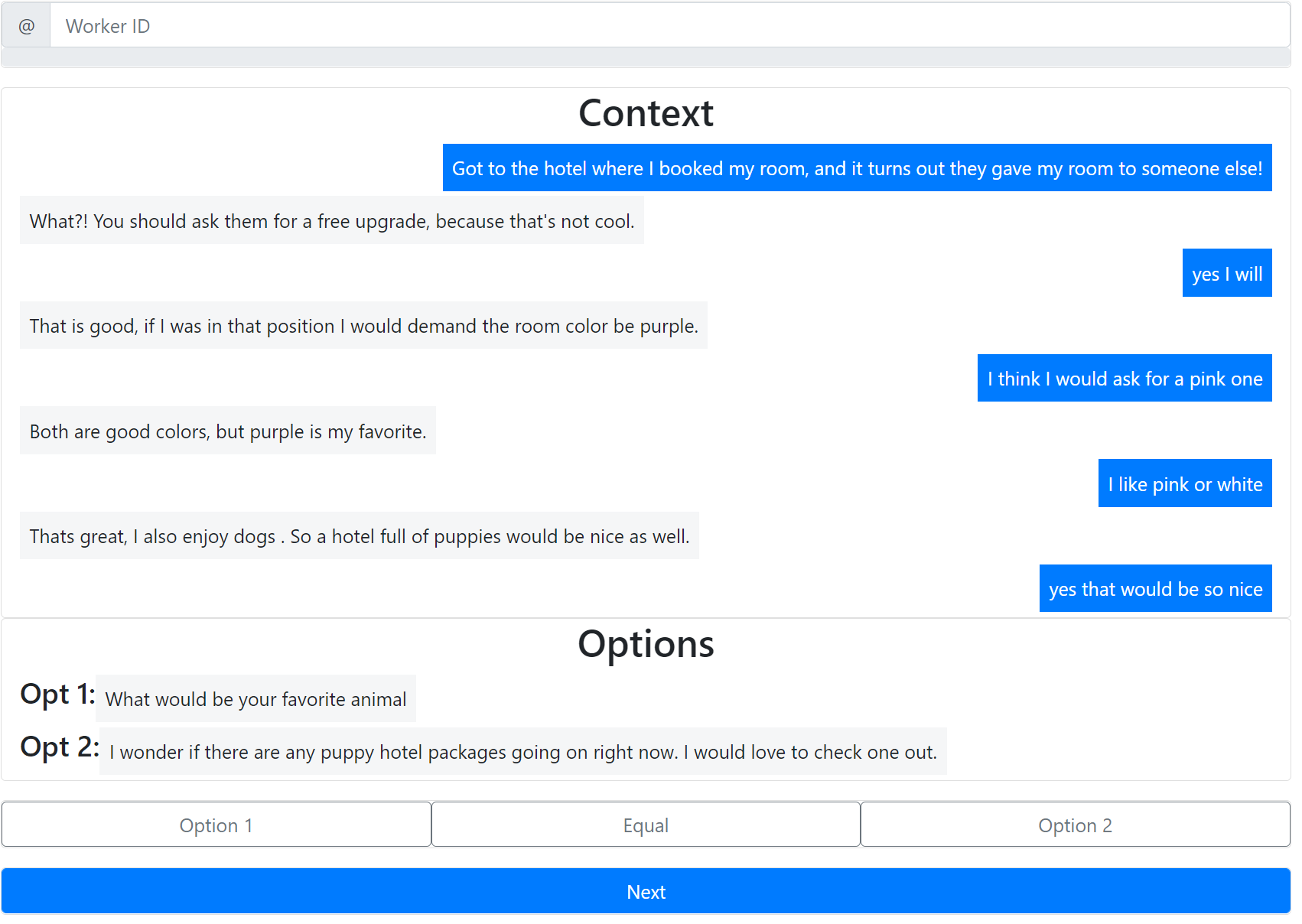}
    \caption{Screenshot of the Dialogue Annotation Tool.}
    \label{fig:dial_ann}
\end{figure*}

\begin{figure}
    \centering
    \includegraphics[width=0.4\textwidth]{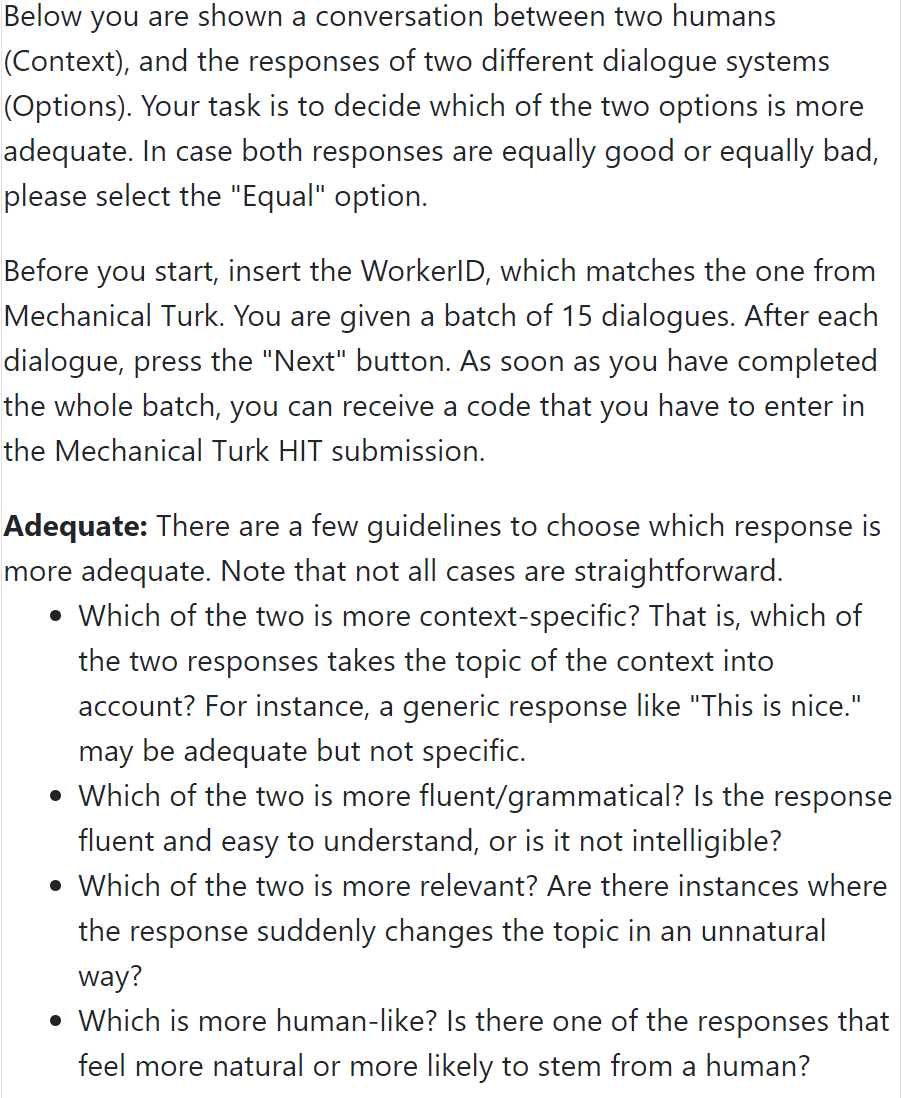}
    \caption{Screenshot of the Instruction of the Dialogue Annotation Tool.}
    \label{fig:dial_ann_instr}
\end{figure}

We let workers on Amazon Mechanical Turk~\footnote{\url{https://www.mturk.com}} perform a preference rating. That is, for each pair of output, the workers decided, which output is more appropriate. Figure~\ref{fig:dial_ann} shows the annotation tool. Each sample was annotated by three workers. Each worker is payed 15 cents per annotation, and at a rate of 1.5 annotations per minute on average, they achieve a wage of 12\$ per hour. We used workers with the Master status and restricted their geographic location to English-speaking countries (USA, UK, Canada, Ireland, and Australia). In Figure~\ref{fig:dial_ann_instr} shows the instructions given to the annotators. The annotations are ratings on the overall adequacy of the utterances. Since each sample is annotated by three different workers, we aggregate the ratings using the MACE~\cite{hovy2013mace} software, which computes a trustworthiness score for each annotator and generates a weighted average to get the final label. Note that this annotation scheme directly yields preference ratings according to our framework.

In order to generate the metric ratings, we used the DialEval framework by~\cite{yeh2021comprehensive}, which integrates a large pool of metrics. We selected five metrics that were easy to setup and achieved decent correlations to human judgments in the evaluation by~\cite{yeh2021comprehensive}. Since the metrics return scalar values for each sample, we create preference ratings as suggested in Definition~\ref{def:derived_metric}. Since the scalar metrics yield real numbered values, there are almost no cases where the derived preference rating yields a draw. 

\subsection{Summarization Data Collection}
For the Summarization domain, we use the data provided by the SummEval framework~\cite{fabbri2021summeval}. It contains data from the Dailymail/CNN dataset~\cite{nallapati2016abstractive}, which contains a test set of 11k samples. The SummEval framework contains the outputs of 23 summarization systems (for a detailed description of these systems, we refer the reader to Section 3.2 of~\cite{fabbri2021summeval}). For 16 of the 23 summarization systems, the authors let 100 generated outputs be rated by three experts on the four characteristics: fluency, consistency, coherence, and relevance. Since these ratings are on a Likert-scale, we transformed those ratings into preference ratings by averaging the three ratings per sample, and applying the transformation proposed in definition~\ref{def:derived_metric}. 

We chose 7 automated metrics based on their popularity and ease of setup. We applied the SummEval framework to generate the automated ratings for each of the 16 summarization systems on the full test set. Analogous to above, we converted the scalar ratings to pairwise ratings by applying definition~\ref{def:derived_metric}. 

\subsection{Machine Translation Data Collection}
For the Machine Translation domain, we used the WMT-21~\cite{freitag2021wmt21metrics} metrics task data. For space limitations, we used the EN$\rightarrow$DE section of the data only. The WMT-21 dataset consists of 15 different metrics for 8 machine translation systems and 3 human references. The human annotations consist of 500 MQM ratings~\cite{lommel2014mqm} done by expert translators. To create preference ratings, we computed the average score of a sample, and compared the score of two outputs of two different systems for the same input by applying definition~\ref{def:derived_metric}. 

The WMT-21 dataset already contains the ratings of the automated metrics for 1000 samples, out of which 500 overlap with the samples annotated by humans. Thus, we generated pairwise ratings by applying definition~\ref{def:derived_metric}.

\section{Full Results}~\label{sec:full_res}
\begin{figure}
\small
\centering
 \begin{subfigure}[c]{0.2\textwidth}
 \centering
     \includegraphics[width=0.55\textwidth]{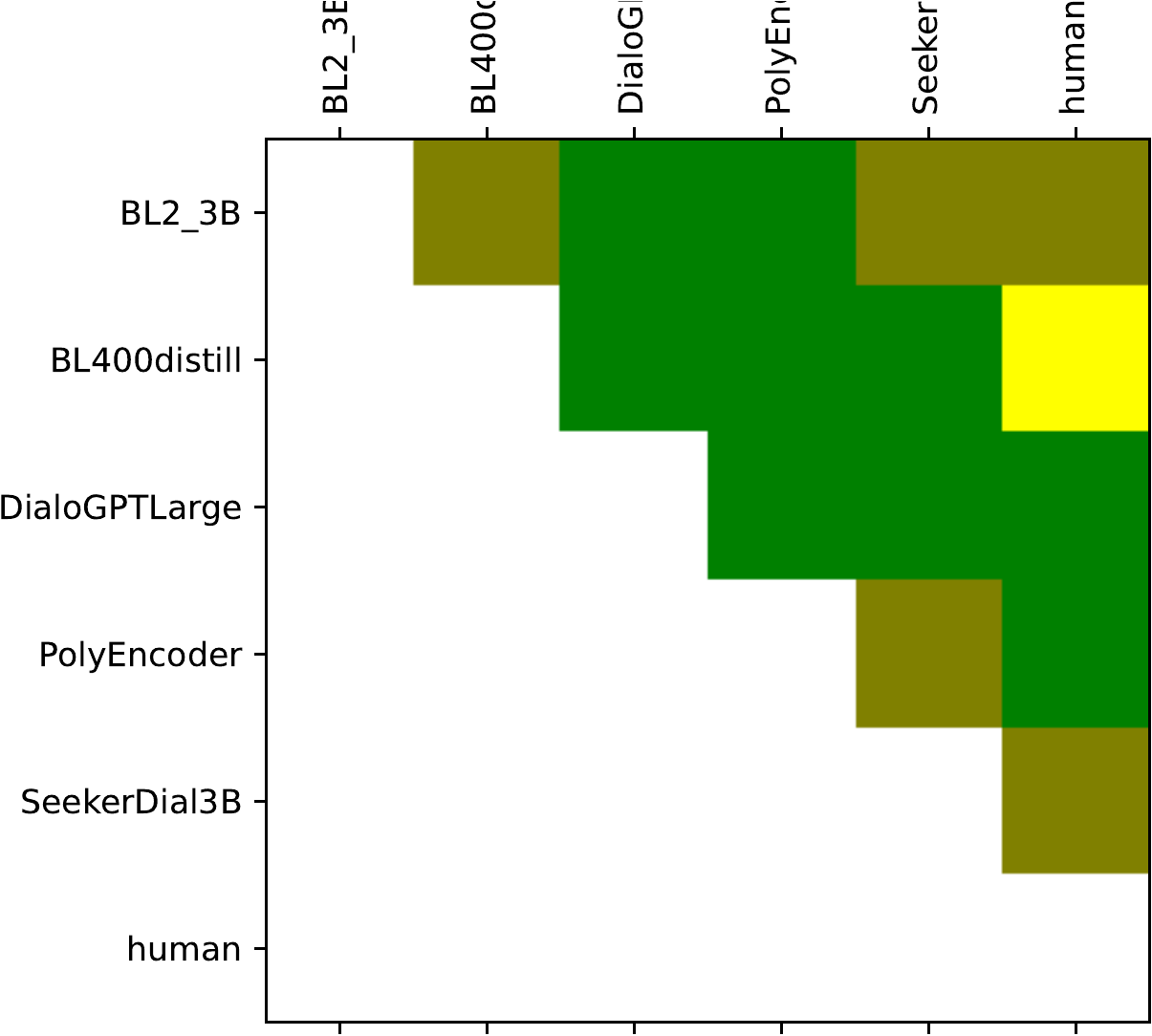}
     \caption{Chatbot DEB}
     \label{fig:cb_deb_naive}
 \end{subfigure}
  \begin{subfigure}[c]{0.2\textwidth}
  \centering
     \includegraphics[width=0.55\textwidth]{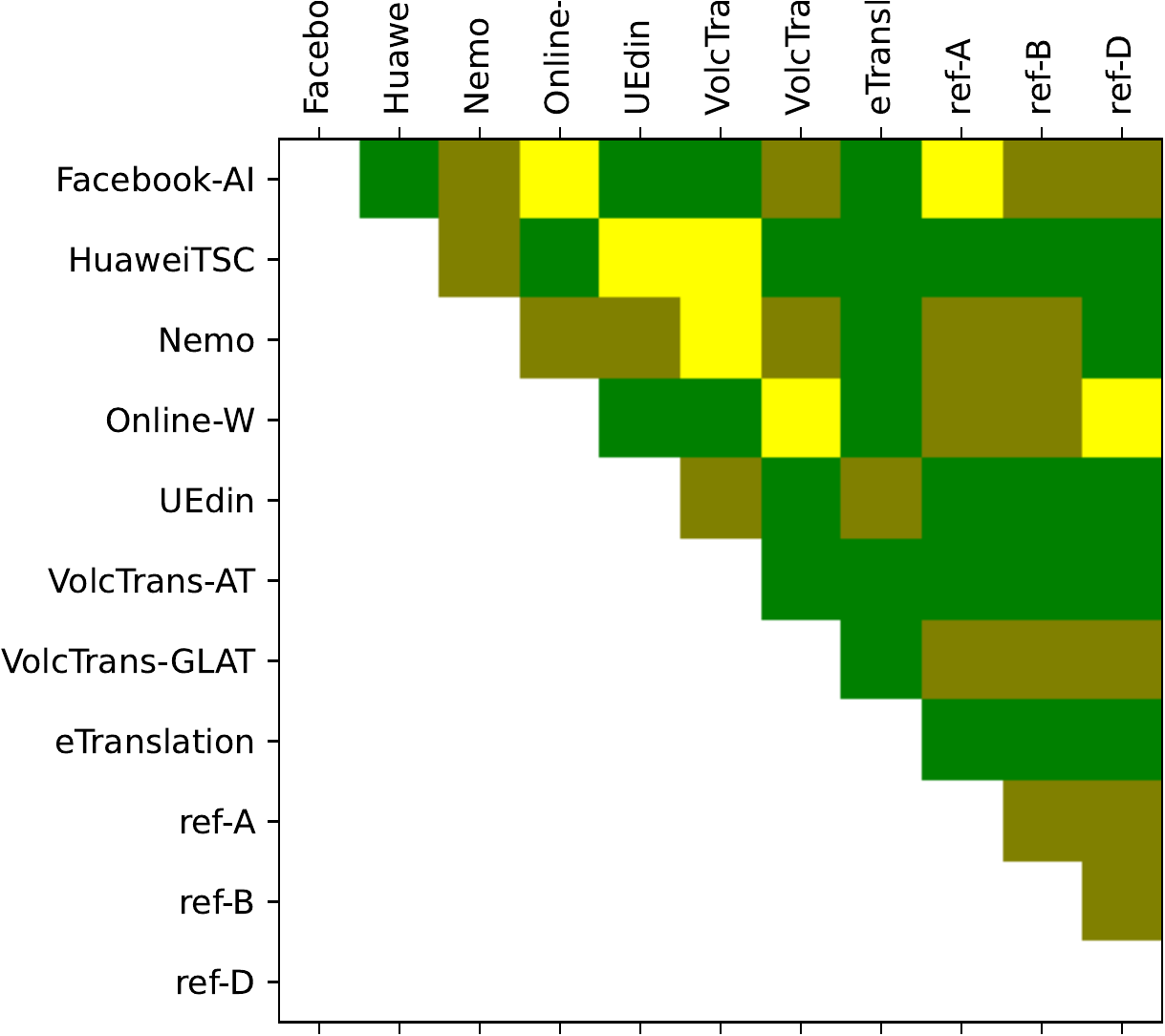}
     \caption{WMT21 COMET}
     \label{fig:wmt_comet_naive}
 \end{subfigure}
 \begin{subfigure}[c]{0.2\textwidth}
 \centering
     \includegraphics[width=0.5\textwidth]{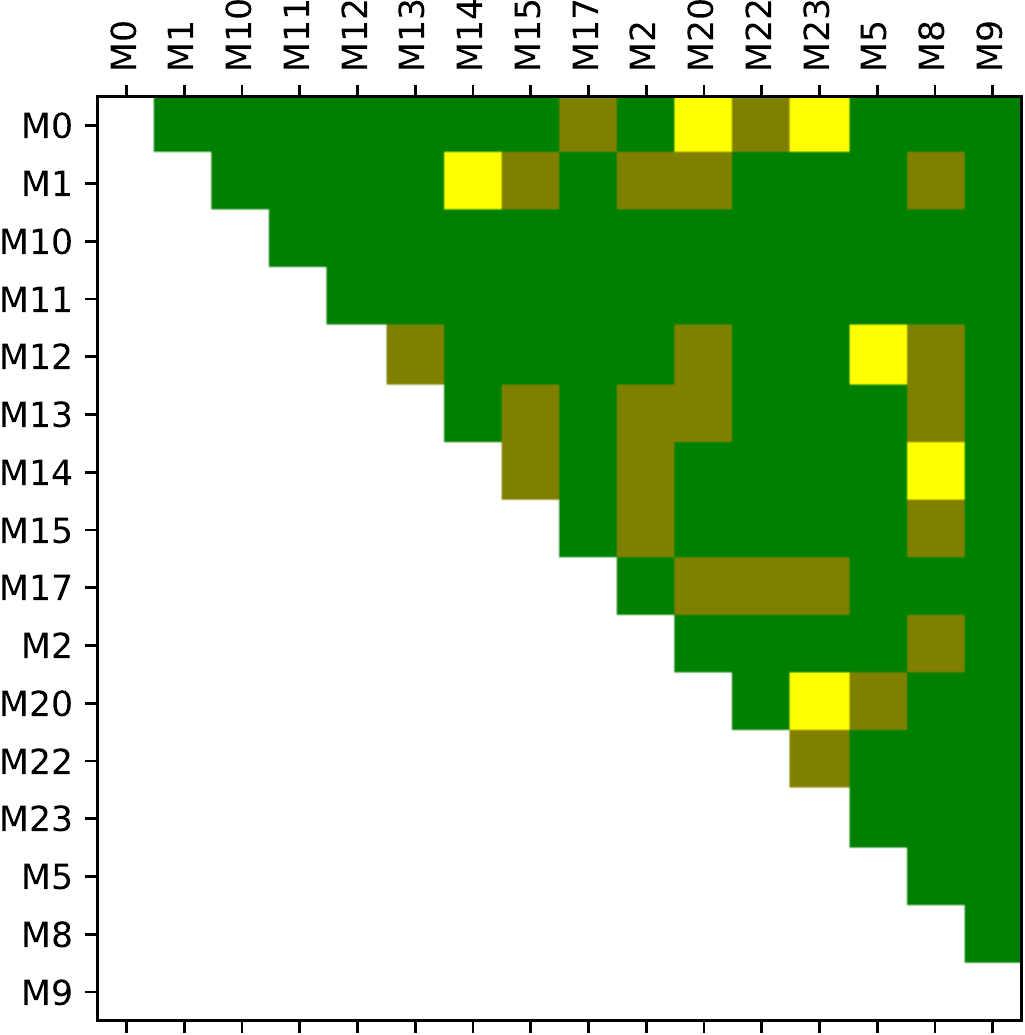}
     \caption{SummEval Coherence Rouge}
     \label{fig:summeval_coh_rouge_naive}
 \end{subfigure}
 \begin{subfigure}[c]{0.2\textwidth}
 \centering
     \includegraphics[width=0.5\textwidth]{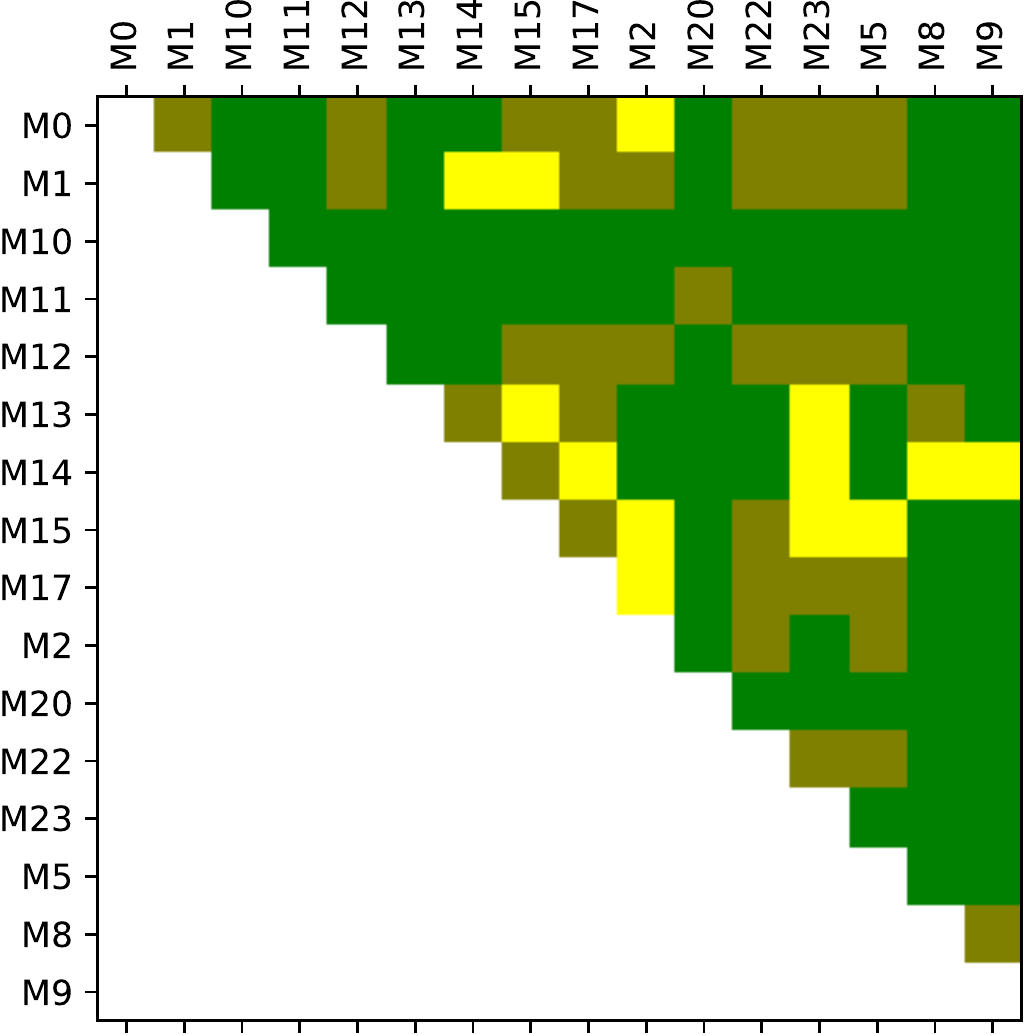}
     \caption{SummEval Consistency BertScore}
     \label{fig:summeval_cons_bert_naive}
 \end{subfigure}

    \caption{Pairwise Errors for the application of automated metrics with the protocol.}
    \label{fig:protocol}
\end{figure}

Tables~\ref{tbl:naive_full_res} and~\ref{tbl:prot_full_res} show the results for all the metrics over all tasks and features. Figure~\ref{fig:protocol} shows the pairwise errors when applying our protocol.

\begin{table}[t!]
\centering
\small
\begin{tabular}{ l | c c c c | c}
Metric & Corr. & Inv. & Omi. & Ins. & KLD \\\hline 
\multicolumn{ 6 }{c}{Chatbot Domain} \\ \hline
deb      & 0.60 & 0.00 & 0.07 & 0.33 & 0.39\\
grade    & 0.53 & 0.07 & 0.20 & 0.20 & 0.53\\
holistic & 0.27 & 0.20 & 0.13 & 0.40 & 0.57\\
maude    & 0.40 & 0.13 & 0.13 & 0.34 & 0.42\\
usl      & 0.53 & 0.07 & 0.00 & 0.40 & 0.40\\ \hline \hline
\multicolumn{ 6 }{c}{Summeval Coherence Domain} \\ \hline
BertScore & 0.58 & 0.22 & 0.00 & 0.20 & 0.35 \\
BLANC     & 0.45 & 0.29 & 0.02 & 0.24 & 0.33 \\
CIDEr     & 0.38 & 0.36 & 0.02 & 0.24 & 0.43 \\
ROUGE-L   & 0.45 & 0.33 & 0.00 & 0.22 & 0.34 \\
S3        & 0.53 & 0.24 & 0.01 & 0.22 & 0.35 \\
SummaQA   & 0.53 & 0.19 & 0.07 & 0.21 & 0.31 \\
SUPERT    & 0.45 & 0.25 & 0.06 & 0.24 & 0.38 \\ \hline \hline
\multicolumn{ 6 }{c}{Summeval Consistency Domain} \\ \hline
BertScore & 0.49 & 0.16 & 0.00 & 0.35 & 1.92 \\
BLANC     & 0.44 & 0.15 & 0.02 & 0.39 & 1.74 \\
CIDEr     & 0.30 & 0.30 & 0.02 & 0.38 & 2.17 \\
ROUGE-L   & 0.35 & 0.25 & 0.02 & 0.38 & 1.98 \\
S3        & 0.46 & 0.17 & 0.01 & 0.37 & 1.90 \\
SummaQA   & 0.55 & 0.09 & 0.03 & 0.33 & 1.81 \\
SUPERT    & 0.46 & 0.13 & 0.04 & 0.38 & 1.76 \\ \hline \hline
\multicolumn{ 6 }{c}{Summeval Fluency Domain} \\ \hline
BertScore & 0.46 & 0.14 & 0.00 & 0.40 & 1.03 \\
BLANC     & 0.40 & 0.16 & 0.01 & 0.43 & 0.98 \\
CIDEr     & 0.34 & 0.21 & 0.02 & 0.43 & 1.08 \\
ROUGE-L   & 0.41 & 0.18 & 0.00 & 0.42 & 0.99 \\
S3        & 0.42 & 0.16 & 0.01 & 0.42 & 1.03 \\
SummaQA   & 0.48 & 0.08 & 0.05 & 0.39 & 0.97 \\
SUPERT    & 0.43 & 0.12 & 0.03 & 0.42 & 1.06 \\ \hline \hline
\multicolumn{ 6 }{c}{Summeval Relevance Domain} \\ \hline
BertScore & 0.64 & 0.13 & 0.01 & 0.22 & 0.26 \\
BLANC     & 0.51 & 0.23 & 0.02 & 0.25 & 0.25 \\
CIDEr     & 0.43 & 0.29 & 0.03 & 0.25 & 0.35 \\
ROUGE-L   & 0.52 & 0.24 & 0.01 & 0.23 & 0.25 \\
S3        & 0.62 & 0.15 & 0.01 & 0.23 & 0.26 \\
SummaQA   & 0.61 & 0.11 & 0.07 & 0.22 & 0.23 \\
SUPERT    & 0.51 & 0.20 & 0.05 & 0.24 & 0.35 \\ 
Ideal-$\mathcal{M}$ & 0.75 & 0.00 & 0.00 & 0.25 & 0.02 \\ \hline \hline
\multicolumn{ 6 }{c}{WMT21 Domain} \\ \hline
BleuRT & 0.47 & 0.02 & 0.13 & 0.38 & 0.46\\
C-SPEC & 0.78 & 0.00 & 0.07 & 0.15 & 0.53\\
COMET  & 0.40 & 0.09 & 0.13 & 0.38 & 0.65\\
BLEU   & 0.44 & 0.16 & 0.07 & 0.33 & 0.43\\ \hline
\end{tabular}

\caption{Frequency of Error Types for all metrics if metrics are applied naively. Correct (Cor.), Inverted (Inv.), Omission (Omi.), Insertion (Ins.), and the fraction of annotations needed with the protocol (Ann.).}
\label{tbl:naive_full_res} 
\end{table}

\begin{table}[t!]
\centering
\small
\begin{tabular}{ l | c c c c | c | c}
Metric & Corr. & Inv. & Omi. & Ins. & KLD & Ann. \\ \hline
\multicolumn{ 7 }{c}{Chatbot Domain} \\ \hline
Human    & 0.93 & 0.00 & 0.00 & 0.07 & 0.05 & 0.59\\
DEB      & 0.93 & 0.00 & 0.00 & 0.07 & 0.06 & 0.49 \\
GRADE    & 0.87 & 0.00 & 0.00 & 0.13 & 0.05 & 0.53 \\
HOLISTIC & 0.87 & 0.00 & 0.00 & 0.13 & 0.05 & 0.52 \\
MAUDE    & 0.87 & 0.00 & 0.00 & 0.13 & 0.05 & 0.53 \\
USL-H    & 0.87 & 0.00 & 0.00 & 0.13 & 0.05 & 0.52 \\ \hline \hline
\multicolumn{ 7 }{c}{Summeval Coherence Domain} \\ \hline
Human     & 0.97 & 0.00 & 0.00 & 0.03 & 0.04 & 0.42 \\
BertScore & 0.96 & 0.00 & 0.00 & 0.04 & 0.04 & 0.38 \\
BLANC     & 0.96 & 0.00 & 0.00 & 0.04 & 0.03 & 0.40 \\
CIDEr     & 0.95 & 0.00 & 0.00 & 0.05 & 0.04 & 0.39 \\
ROUGE-L   & 0.96 & 0.00 & 0.00 & 0.04 & 0.03 & 0.40 \\
S3        & 0.97 & 0.00 & 0.00 & 0.03 & 0.03 & 0.41 \\
SummaQA   & 0.94 & 0.00 & 0.01 & 0.05 & 0.04 & 0.39 \\
SUPERT    & 0.94 & 0.00 & 0.00 & 0.06 & 0.04 & 0.38 \\ \hline \hline
\multicolumn{ 7 }{c}{Summeval Consistency Domain} \\ \hline
Human     & 0.93 & 0.00 & 0.00 & 0.07 & 0.02 & 0.53 \\
BertScore & 0.88 & 0.00 & 0.00 & 0.12 & 0.02 & 0.45 \\
BLANC     & 0.88 & 0.00 & 0.00 & 0.12 & 0.02 & 0.48 \\
CIDEr     & 0.87 & 0.00 & 0.01 & 0.13 & 0.02 & 0.46 \\
ROUGE-L   & 0.88 & 0.00 & 0.00 & 0.12 & 0.02 & 0.46 \\
S3        & 0.88 & 0.00 & 0.00 & 0.12 & 0.02 & 0.46 \\
SummaQA   & 0.90 & 0.00 & 0.00 & 0.10 & 0.02 & 0.47 \\
SUPERT    & 0.88 & 0.00 & 0.00 & 0.12 & 0.02 & 0.46 \\ \hline \hline
\multicolumn{ 7 }{c}{Summeval Fluency Domain} \\ \hline
Human     & 0.95 & 0.00 & 0.00 & 0.05 & 0.03 & 0.60 \\
BertScore & 0.91 & 0.00 & 0.01 & 0.08 & 0.04 & 0.56 \\
BLANC     & 0.91 & 0.00 & 0.01 & 0.08 & 0.02 & 0.58 \\
CIDEr     & 0.93 & 0.00 & 0.00 & 0.07 & 0.04 & 0.55 \\
ROUGE-L   & 0.93 & 0.00 & 0.00 & 0.07 & 0.04 & 0.57 \\
S3        & 0.92 & 0.00 & 0.01 & 0.08 & 0.04 & 0.57 \\
SummaQA   & 0.92 & 0.00 & 0.01 & 0.08 & 0.03 & 0.58 \\
SUPERT    & 0.88 & 0.00 & 0.01 & 0.11 & 0.04 & 0.54 \\ \hline \hline
\multicolumn{ 7 }{c}{Summeval Relevance Domain} \\ \hline
Human     & 0.95 & 0.00 & 0.00 & 0.05 & 0.07 & 0.43 \\
BertScore & 0.95 & 0.00 & 0.01 & 0.04 & 0.06 & 0.40 \\
BLANC     & 0.90 & 0.00 & 0.03 & 0.07 & 0.05 & 0.42 \\
CIDEr     & 0.94 & 0.00 & 0.02 & 0.04 & 0.06 & 0.42 \\
ROUGE-L   & 0.94 & 0.00 & 0.01 & 0.05 & 0.05 & 0.41 \\
S3        & 0.95 & 0.00 & 0.01 & 0.04 & 0.06 & 0.41 \\
SummaQA   & 0.93 & 0.00 & 0.01 & 0.06 & 0.05 & 0.41 \\
SUPERT    & 0.93 & 0.00 & 0.02 & 0.05 & 0.06 & 0.39 \\ 
Ideal-$\mathcal{M}$ & 0.85 & 0.00 & 0.00 & 0.15 & 0.02 & 0.38 \\ \hline \hline
\multicolumn{ 7 }{c}{WMT21 Domain} \\ \hline
Human      & 0.65 & 0.00 & 0.00 & 0.35 & 0.06 & 0.34\\
BleuRT     & 0.65 & 0.00 & 0.00 & 0.35 & 0.06 & 0.34 \\
C-SPEC     & 0.65 & 0.00 & 0.00 & 0.35 & 0.06 & 0.33 \\
COMET      & 0.65 & 0.00 & 0.00 & 0.35 & 0.06 & 0.34 \\
BLEU       & 0.65 & 0.00 & 0.00 & 0.35 & 0.05 & 0.34 \\ \hline
\end{tabular}

\caption{Frequency of Error Types for all metrics if the protocol is applied. Correct (Cor.), Inverted (Inv.), Omission (Omi.), Insertion (Ins.), and the fraction of annotations needed with the protocol (Ann.).}
\label{tbl:prot_full_res} 
\end{table}

% \begin{figure}
% \small
% \centering
%      \begin{subfigure}[c]{0.2\textwidth}
%      \centering
%          \includegraphics[width=0.55\textwidth]{figures/chatbot_deb_corrected.pdf}
%          \caption{Chatbot DEB}
%          \label{fig:cb_deb_naive}
%      \end{subfigure}
%      \begin{subfigure}[c]{0.2\textwidth}
%           \centering
%          \includegraphics[width=0.55\textwidth]{figures/wmt21_comet_corrected.pdf}
%          \caption{WMT21 COMET}
%          \label{fig:wmt_comet_naive}
%      \end{subfigure}
%      \begin{subfigure}[c]{0.2\textwidth}
%      \centering
%          \includegraphics[width=0.5\textwidth]{figures/summeval_coherence_rouge_corrected.pdf}
%          \caption{SummEval Coherence Rouge}
%          \label{fig:summeval_coh_rouge_naive}
%      \end{subfigure}
%      \begin{subfigure}[c]{0.2\textwidth}
%      \centering
%          \includegraphics[width=0.5\textwidth]{figures/summeval_consistency_bert_score_corrected.pdf}
%          \caption{SummEval Consistency BertScore}
%          \label{fig:summeval_cons_bert_naive}
%      \end{subfigure}

%         \caption{Correction with uniform prior of Automated Metrics.}
%         \label{fig:corrected}
% \end{figure}

\section{Ideal Metric}~\label{sec:ideal_met}
Since the real-world metrics that we use in this work are not yet of very high quality, we also generated synthetic data for the SummEval-Relevance domain. That is, we simulated a metric using a fixed $\bm{\mu}_{sim} = \big(\begin{smallmatrix}
            0.8 & 0.25 & 0.1 \\
            0.1 & 0.5 & 0.1 \\
            0.1 & 0.25 & 0.8 \\
        \end{smallmatrix}\big)$. For this, we used the human win-rates $\bm{p}$ to randomly assign labels for each pair of system for the 11k samples. That is, for each sample, we randomly chose rating $r~\bm{p}$. These sampled labels are treated as the ground-truth, which has the same distribution as the human labels. Then we used $\bm{\hat{p}} = \bm{\mu}_{sim}\bm{p}$ to generate corrupted labels, which correspond to metric ratings. Since $\bm{\mu}_{sim}$ makes less mistakes than the existing metrics, and it makes no Omission or Inversion Errors. We added the ideal metric to the full results in Table~\ref{tbl:naive_full_res} and~\ref{tbl:prot_full_res}. It also achieves a very good KLD score when applied naively. However, such a metric is currently out of scope and this only serves to illustrate an upper bound for automated metrics. 
\end{document}